\documentclass[final,5p,times]{elsarticle}

\usepackage{amssymb}
\usepackage{graphicx}
\usepackage{amsmath}
\usepackage[utf8]{inputenc}
\usepackage[small]{caption}
\usepackage{times}
\usepackage{soul}
\usepackage{url}
\usepackage{booktabs}
\usepackage{algorithm}
\usepackage{algorithmic}
\usepackage{latexsym}
\usepackage{multirow}
\usepackage{makecell}
\usepackage{color}
\usepackage{array}
\usepackage[figuresright]{rotating}
\usepackage{pdflscape}
\definecolor{ddg}{rgb}{0.48,0.48,0.48}
\definecolor{dg}{rgb}{0.58,0.58,0.58}
\definecolor{g}{rgb}{0.68,0.68,0.68}
\definecolor{lg}{rgb}{0.78,0.78,0.78}
\definecolor{llg}{rgb}{0.9,0.9,0.9}

\journal{}

\begin{document}

\begin{frontmatter}



\title{Modeling Relation Paths for Knowledge Base Completion via Joint Adversarial Training}

\author[label1,label2]{Chen Li}
\author[label1,label2]{Xutan Peng}
\author[label3]{Shanghang Zhang}
\author[label1,label2]{Hao Peng}
\author[label4]{Philip S. Yu}
\author[label5]{Min He}
\author[label1]{Linfeng Du}
\author[label5]{Lihong Wang*}
\address[label1]{Beijing Advanced Innovation Center for Big Data and Brain Computing, Beihang University, Beijing, China}
\address[label2]{State Key Laboratory of Software Development Environment Lab, Beihang University, Beijing, China}
\address[label3]{Department of Electrical and Computer Engineering, Carnegie Mellon University, Pittsburgh, CP, USA}
\address[label4]{Department of Computer Science, University of Illinois at Chicago, Chicago, IL, USA}
\address[label5]{National Computer Network Emergency Response Technical Team/Coordination Center of China, Beijing, China}



\begin{abstract}
Knowledge Base Completion (KBC), which aims at determining the missing relations between entity pairs, has received increasing attention in recent years.
Most existing KBC methods focus on either embedding the Knowledge Base (KB) into a specific semantic space or leveraging the joint probability of Random Walks (RWs) on multi-hop paths.
Only a few unified models take both semantic and path-related features into consideration with adequacy.
In this paper, we propose a novel method to explore the intrinsic relationship between the single relation (i.e. 1-hop path) and multi-hop paths between paired entities.
We use Hierarchical Attention Networks (HANs) to select important relations in multi-hop paths and encode them into low-dimensional vectors.
By treating relations and multi-hop paths as two different input sources, we use a feature extractor, which is shared by two downstream components (i.e. relation classifier and source discriminator), to capture shared/similar information between them.
By joint adversarial training, we encourage our model to extract features from the multi-hop paths which are representative for relation completion.
We apply the trained model (except for the source discriminator) to several large-scale KBs for relation completion. Experimental results show that our method outperforms existing path information-based approaches.
Since each sub-module of our model can be well interpreted, our model can be applied to a large number of relation learning tasks.
\end{abstract}


\end{frontmatter}


\section{Introduction}\label{s1}
Knowledge Base, which is used to store structured knowledge data, has achieved satisfactory results in many artificial intelligence tasks such as question answering~\cite{Fader2014kdd}, information retrieval~\cite{Nikolaev2016sigir}, and relation extraction~\cite{relationkg2018hci}.
Existing large-scale KBs (e.g., Freebase~\cite{Bollacker2008kdd} and WordNet~\cite{Miller1995acm}) contain a vast amount of facts about the real world in the form of triples such as (\textit{Head Entity}, \texttt{Relation}, \textit{Tail Entity}).
However, the growth of edges in KBs is not in proportion to that of entities due to the lack of data collection~\cite{Bordes2011aaai}.
In other words, edges in existing KBs are extremely sparse and far from complete.
Complementing most of the missing facts manually will cost an unacceptable amount of time and resources.
Hence, a large number of Knowledge Base Completion techniques based on analyzing the labeled data have been proposed.

The existing KBC methods can be roughly divided into two types: Knowledge Representation (KR) based methods, and Path Ranking (PR) based methods.
KR-based methods, represented by the translation models~\cite{Bordes2013nips, Wang2014aaai, Lin2015aaai}, which embed the knowledge in a KB into a specific continuous vector space and exploit uniform spatial relationships in the projective space to describe the corresponding relations.
Based on the representation of the given entity pair, KR-based methods define a score function to rank candidate relations to complete the KB.
Meanwhile, there exist Path Ranking (PR) based methods, which enumerate and select valuable paths among the given entity pair as the features to be analyzed.
PR-based methods determine the missing relation through training Random Walk (RW) joint probability of selected paths, and also the corresponding relation classifier~\cite{Lao2010ML}.
PR-based methods and their variants have achieved a satisfactory result on KB completion by optimizing the calculation strategies of path similarity and path selection~\cite{Lao2015acl, Mazumder2017ijcai}.

\begin{figure}[t]
	\centering
	\includegraphics[width=0.47\textwidth]{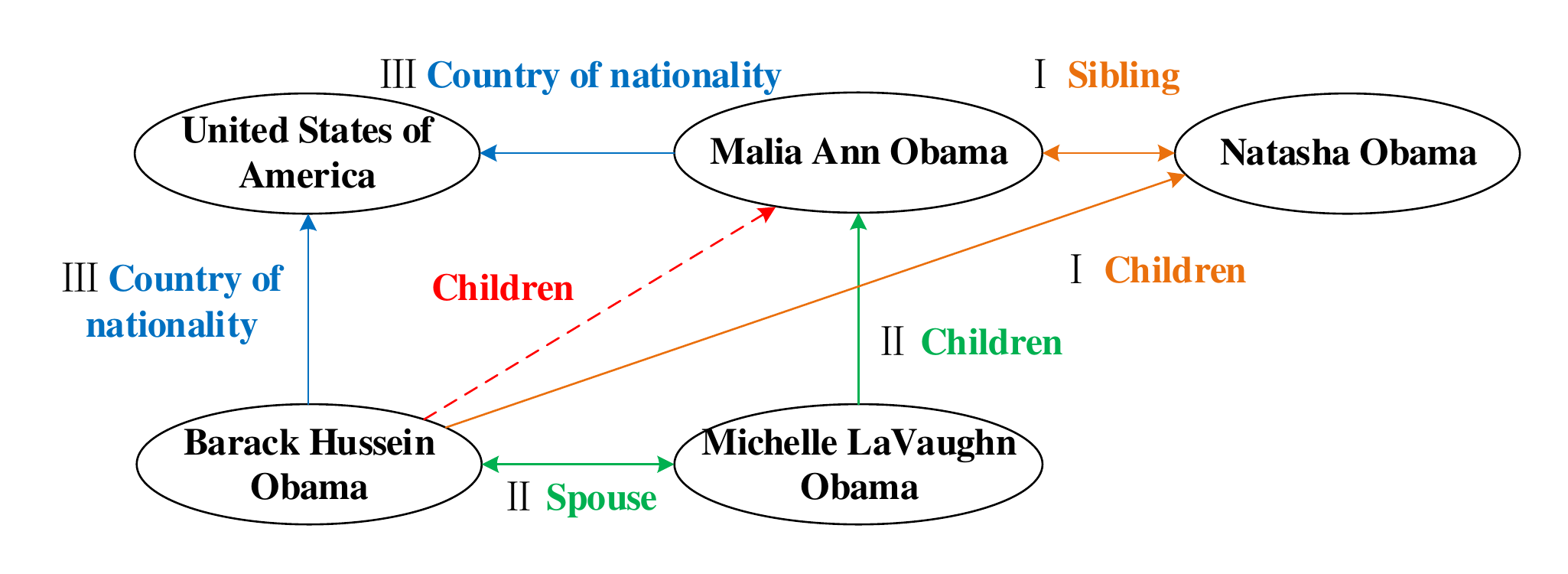}
	\caption{
		An example of KBC in Freebase. 
		The missing single relation is \texttt{Children} (which can be indicated by the red dotted line).
	}\label{fig:example}
	\vspace{-0.5cm}
\end{figure}

A drawback of the KR-based models is that they only take semantic information implied by the single relations (1-hop paths) into consideration, thus ignoring the interpretation of multi-hop paths among the paired entities~\cite{lin2015emnlp,shen2016arxiv}.
For example, as shown in Figure~\ref{fig:example}, the missing relation \texttt{Children} can be inferred by observing the multi-hop paths between \textit{Barack Hussein Obama} and \textit{Malia Ann Obama} (i.e. paths I (yellow) and II (green)).
More generally, the semantic relationship and co-occurrence information is intrinsically built in the single relation and multi-hop paths between the paired entities.
Based on this principle, PR-based methods capture the features of the multi-hop paths and calculate the joint probability of each candidate relation.
However, PR-based approaches also have their drawbacks, which lie in the existence of noise within multi-hop paths (as illustrated in Figure~\ref{fig:example}, path III (blue) with two hops is useless for determining the missing relation) as well as the lack of mining the inner relationships between single relation and multi-hop paths~\cite{Mazumder2017ijcai}.

Based on an intuitive assumption that a vast amount of paths between the paired entities generally contain shared/similar semantic and co-occurrence information, this paper presents a unified model that takes both the single relation and multi-hop paths between the given entity pair into full account.
Specifically, our model complements the missing single relation by extracting shared information from multi-hop paths.
To achieve that, our model leverages Hierarchical Attention Networks (HANs) to encode the inputs and send the features to a downstream feature extraction module, where shared features are automatically captured without any manual selection through Adversarial Training (AT).
Among them, the HANs is used to ensure that the features fed to the downstream modules retain the important information in multi-hop paths.
The feature extraction module consists of two tightly coupled and parameter-shared neural network structures, one for relation prediction and one for source discrimination.
The whole framework is jointly adversarial trained so that the extracted features not only minimize the classification error but also make the source discriminator unable to discriminate their source.
Finally, we can get the partial information of a single relation by only taking multi-hop paths as input, so as to determine the missing relation.

The experimental results on FB and WN demonstrate that our method outperforms the state-of-the-art approaches.
More generally, since the sub-modules in our model can be well interpreted, our method can transform various tasks (e.g., reasoning and information extraction) into encoding and decoding tasks by defining appropriate adversarial data sources.

The contributions of this paper can be summarized as follows:
\begin{itemize}
	\item We propose a general approach to mine the intrinsic relationship between single relation and multi-hop paths based on specific downstream tasks.
	\item We propose a novel KBC approach which outperforms the state-of-the-art methods.
	\item By visualizing and analyzing the working processes of the components in our model, we interpret their corresponding functions and effects.
\end{itemize}

The rest of this paper is organized as follows:
Section~\ref{s2} reviews traditional KBC methods and the AT framework.
Section~\ref{s3} details the composition of the whole framework and each of the sub-modules proposed in this paper, and conducts a theoretical analysis.
Section~\ref{s4} introduces the experimental details and analyzes the results of typical downstream tasks.
At length, section~\ref{s5} concludes this paper.

\section{Related Work}\label{s2}

\subsection{Knowledge Base Completion}
As mentioned above, the existing methods for KBC mainly include the KR-based models~\cite{Bordes2011aaai,Bordes2013nips,Bordes2014ML,Wang2014aaai,Lin2015aaai} and PR-based models~\cite{Lao2010ML,Gardner2014emnlp,Lao2015acl,Mazumder2017ijcai}.

KR-based models, originating from representation learning, focus on modeling triples in structured~\cite{Bordes2011aaai} or unstructured~\cite{Bordes2014ML} embeddings.
Trans-family models~\cite{Bordes2013nips, Wang2014aaai, Lin2015aaai} train embeddings based on the assumed spatial positional relationship that $\textbf{h}+\textbf{r}\approx\textbf{t}$.
Recent years have seen more elaborate triple modeling methods including:
DistMult, which defines a triple scoring method based on matrix operation~\cite{distmult2015iclr};
ComplEx, which introduces complex space into KR to handle two-way relations~\cite{complex2016icml}.
Also, The work of Sun et al.~\cite{rotate2019axriv} focuses on spatial rotation to optimize triple modeling.
In addition, ConvE improves the quality of KR by modeling the relations through convolution methods~\cite{conv2d2018aaai}.
Similarly, Shang et al.~\cite{cnnkbc2019aaai} and Nguyen et al.~\cite{cnnkbc2018naacl} achieve KBC based on convolutional neural networks.
Nguyen et al.~\cite{capsule2019naacl} introduce capsule neural embedding architecture for learning knowledge embedding to complement the relations.
MultiE models triples based on multi-task learning~\cite{multitask2018cikm}.
Moreover, through utilizing canonical tensor decomposition, Lacroix et al.~\cite{decompostion2018icml} obtain high-quality knowledge embedding for KBC.
In general, the KR-based methods vectorize the entities and relations and achieve the purpose of KBC by ranking candidate relations.

In contrast to KR-based methods, PR-based methods generally leverage a RW-based inference technique.
Initially proposed by Lao and Cohen~\cite{Lao2010ML}, PR-based methods determine the missing edges by planning the optimal RW path between entities, and calculating the joint probabilities of the selected paths.
Further innovations lay in variant methods which can be summarized as the contextual information/feature extraction of paths~\cite{Mazumder2017ijcai}, the planning of RW~\cite{Lao2015acl}, and the optimization of path similarity calculation~\cite{Gardner2014emnlp}.

Recently, some research has begun focusing on incorporating extra information into the completion model.
Among them, Lin et al.~\cite{lin2015emnlp} introduce low-order (2 or 3) path-related information into TransE in three different ways (ADD, MUL, and RNN) and optimize the model.
Dur{\'{a}}n et al.~\cite{Garcdur2015emnlp} propose an extension of TransE that learns to explicitly model the composition of relations via the addition of their corresponding translation vectors.
The work of Xie et al.~\cite{xie2017acl} equips the Trans-family model with a sparse attention, which represents the hidden concepts of relations and transfers statistical strength.
Wei et al.~\cite{wei2016emnlp} combine external memory and the RW strategy to mine the inference formulas.
The method proposed by Shen et al.~\cite{shen2016arxiv} uses a global memory and a controller module to learn multi-hop paths in vector space and infers missing facts jointly without any human-designed procedure.
Yoshikawa et al.~\cite{axiom2019aaai} achieve KBC by combining axiom injection as an inference basis.                                                   
By constructing rich features, Komninos and Manandhar~\cite{featurerich2017acl} use a neural network to complete missing relations.

Table~\ref{tab:complexity} lists the time and space complexity of some classical methods.
Within the experimental setting of 3 as the upper limit of the paths' hop number, we can conclude that the time complexity of our model on the FB series datasets is less than that of the other methods', and the time complexity on the WN series datasets is close to $\mathcal{O}(2N_{t})$.
With improvements in time complexity, our model's space complexity is similar to that of the others'.

\begin{table}
	\centering
	\caption{Complexity (the number of parameters and the number of multiplication operations in an epoch)
		of several models.
		$N_{e}$ and $N_{r}$ represent the number of entities and relations, respectively.
		$N_{t}$ represents the number of triplets in a knowledge graph. 
		$N_{p,i}$ denotes the number of i-hop paths in dataset.
		$m$ is the dimension of entity embedding space and $n$ is the dimension of relation embedding space. 
		$d$ denotes the average number of clusters of a relation. 
		$l_{i}$ is the length of i-hop paths.
		$k$ is the number of hidden nodes of a neural network and s is the number of slices per tensor.
		In both datasets of FB15K* and WN18* $N_{p,i} < N_{t}$ with $N_{t}/N_{p,i}=$ 15.7 and 1.97 respectively.
	}
	\label{tab:complexity}
	\begin{tabular}{l|c|c}
		\hline
		 & Time Complexity & Space Complexity \\
		\hline
		TransE & $\mathcal{O}(N_{t})$ & $\mathcal{O}(N_{e}m+N_{r}n)$ \\
		TransH & $\mathcal{O}(2mN_{t})$ & $\mathcal{O}(N_{e}m+2N_{r}n)$ \\
		TransR & $\mathcal{O}(2mnN_{t})$ & $\mathcal{O}(N_{e}m+N_{r}(m+1)n)$ \\
		PTransE(ADD) & $\mathcal{O}(N_{t})$ & $\mathcal{O}(N_{e}m+N_{r}n)$ \\
		PTransE(MUL) & $\mathcal{O}(N_{t}+\sum l_{i}N_{p,i})$ & $\mathcal{O}(N_{e}m+N_{r}n)$ \\
		PTransE(RNN) & $\mathcal{O}(N_{t}+\sum l_{i}N_{p,i})$ & $\mathcal{O}(N_{e}m+N_{r}n)$ \\
		\hline
		DistMult & $\mathcal{O}(2N_{t})$ & $\mathcal{O}(N_{e}m+N_{r}n)$ \\
		ComplEx & $\mathcal{O}(2N_{t})$ & $\mathcal{O}(N_{e}m+N_{r}n)$ \\
		ConvE & $\mathcal{O}(n'N_{t})$ & $\mathcal{O}(N_{e}m+N_{r}n')$ \\
		RotatE & $\mathcal{O}(2N_{t})$ & $\mathcal{O}(N_{e}m+N_{r}n)$ \\
		\hline
		Ours & $\mathcal{O}(\sum (l_{i}+1)N_{p,i})$ & $\mathcal{O}(N_{e}m+N_{r}n)$ \\
		\hline
	\end{tabular}
\end{table}

The common drawback of the existing models is the lack of mining the internal semantic relations between single relation and multi-hop paths, which could bring in path-related information for KBC.
In this paper, we present a novel method to exploit the intrinsic semantic and co-occurrence relationships between single relation and multi-hop paths, and use them to complete the knowledge base.

\begin{figure*}[ht]
	\centering
	\includegraphics[width=1\textwidth]{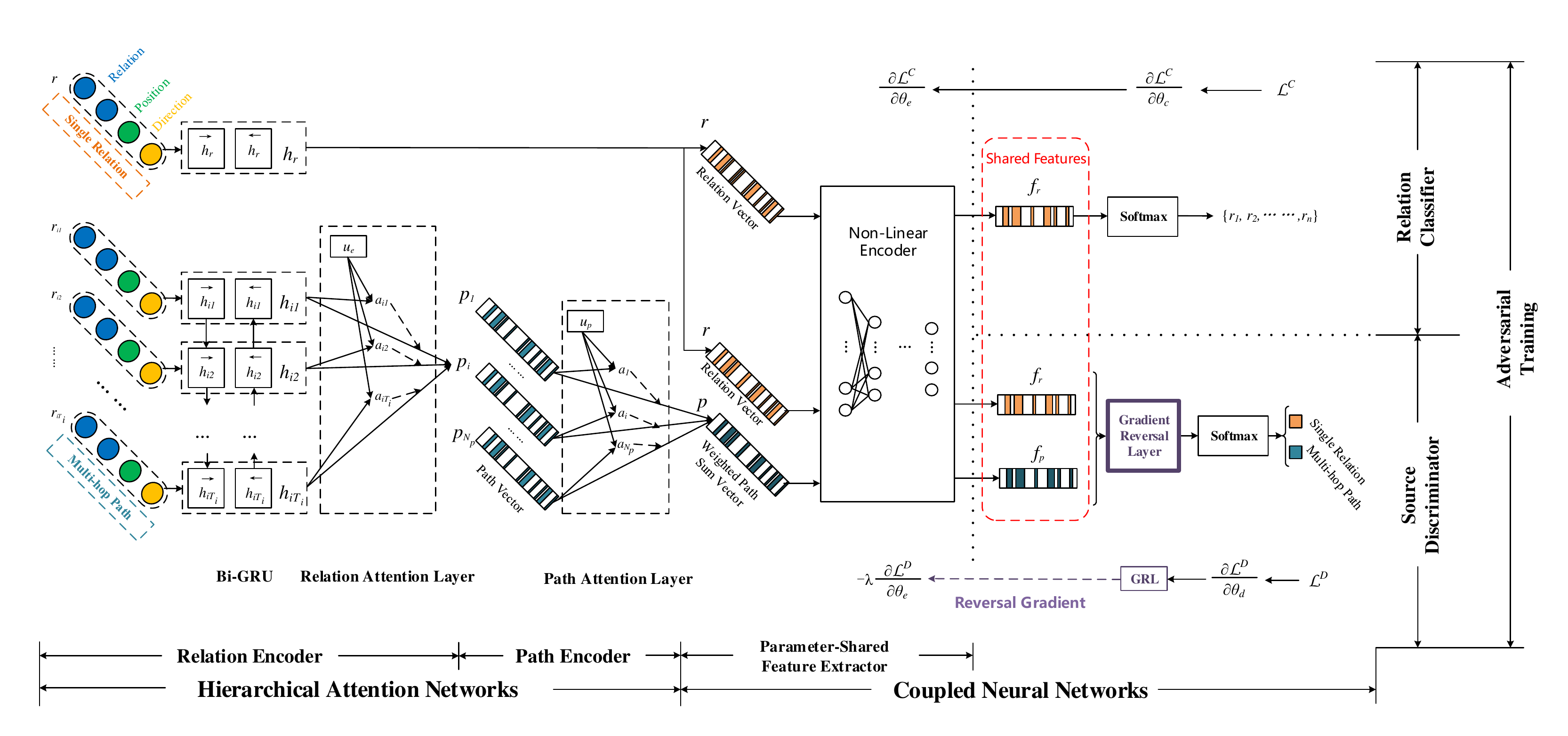}
	\caption{The overview of the architecture of the proposed model.}\label{fig:overall}
\end{figure*}

\subsection{Adversarial Training}
Inspired by the two player game, the Generative Adversarial Networks (GANs) consist of two connected sub-modules, in which the discriminator is trained to distinguish if the input sample matches the true data distribution or is generated by the generator while the generator is trained to generate fake samples that maximize the error rate of the discriminator.
GANs have been successful in a large number of image-related tasks, and the adversarial framework in GANs has been adapted to fit many other tasks by introducing additional structures.
Among them, Han et al.~\cite{disre2018axriv} propose a method for learning implicit features by merging the adversarial mechanism into auto-encoder, which realizes the processing of general data.
Tadesse and Collons~\cite{atwe2018aaai} utilize the adversarial framework to unify the semantic information of different sources into word-char embeddings, and have achieved improvements in tasks of sequence labeling.
The work of Qin et al.~\cite{dsgan2018acl} integrates the adversarial mechanism into the screening of training samples, which improves the quality of input samples of distant supervision, and consequently improves the performance of information extraction.
Yasunaga et al.~\cite{POSat2018naacl} analyze the specific effects of the adversarial mechanism in natural language processing in detail and propose a robust model for multi-lingual part-of-speech tagging tasks.
In summary, we find that AT can extract shared information between different sources effectively under specific requirements.

\section{Methodology}\label{s3}
In this section, we introduce the proposed KBC method in detail.

\subsection{Problem Definition and Notations}
Given a KB containing a collection of triples $\mathcal{T}=\{(h,r,t) \mid h,t\in E,r\in R\}$, where $h,t$ respectively denote the head and tail entity, $r$ denotes the single relation.
$E$ and $R$ are the entity and relation sets.
A multi-hop path is a sequence of relations $p_{i}=(r_{i1},r_{i2},\dots,r_{iT_{i}}), p_{i} \in \mathcal{P}, r_{ij} \in R, i \in [1,N_{p}], j \in [1,T_{i}]$, where $p_{i}$ is a path in the multi-hop paths set $\mathcal{P}$, and $N_{p}$ and $T_{i}$ denote the size of $\mathcal{P}$ and the number of relations in $p_{i}$.
$\mathcal{P}$ contains all of the paths between $(h,t)$ except the single relation $r$.
The task of KBC is to train a relation classifier based on the labeled data $\{r\} \bigcup \mathcal{P}$ and complement the missing relation $r$ between the given entity pair $(h,t)$.

\subsection{Overall Architecture and Components}
The overall architecture of the proposed method is shown in Figure~\ref{fig:overall}.
The hierarchical attention-based encoders are used to encode the single relation and multi-hop paths as input, and the coupled neural networks are used to extract features.
We will describe the detail of different components in the following sections.

\subsubsection{Hierarchical Attention-Based Encoders}
Given a path $p_{i},i \in [1,N_{p}]$, we first embed the relations into vectors through an embedding matrix $W_{t}, v_{ij}=W_{t}r_{ij}$, where $r_{ij}$ denotes the $j$th hop relation in $p_{i}$.
Additionally, we add Position Encoding~\cite{Sukhbaatar2015nips} $pe_{ij}$ and direction information $d_{ij}$ after the relation representation to get the final input vector $x_{ij}=[v_{ij}; pe_{ij}; d_{ij}]$.
We utilize a Bidirectional Gated Recurrent Unit (Bi-GRU)~\cite{Bahdanau2014axriv} to get the annotations of the relation sequence, which summarizes the information from both directions.

As illustrated in the left part of Figure~\ref{fig:overall}, we obtain the annotation for each relation $r_{ij}$ in $p_{i}$ by concatenating the forward hidden state $\overrightarrow{h_{ij}}$ and the backward hidden state $\overleftarrow{h_{ij}}$, i.e. $h_{ij}=[\overrightarrow{h_{ij}};\overleftarrow{h_{ij}}]$.
Among them, $\overrightarrow{h_{ij}}$ (or $\overleftarrow{h_{ij}}$) is computed as
\begin{small}
	\begin{gather*}
		\overrightarrow{h_{ij}}=(1-z_{ij})\odot h_{i,j-1}+z_{ij}\odot \widetilde{h}_{ij},\tag{1}
	\end{gather*}
\end{small}
where $\overrightarrow{h_{ij}}$ is a linear interpolation between the previous state $h_{i,j-1}$ and the candidate state $\widetilde{h}_{ij}$.

GRU utilizes a gating mechanism to track the states of the sequence.
Update gate $z_{ij}$ decides how much past information is kept and how much new information is added, which is computed as
\begin{small}
	\begin{gather*}
	z_{ij}=\sigma(W_{z}x_{ij}+U_{z}h_{i,j-1}+b_{z})\tag{2},
	\end{gather*}
\end{small}
where $\sigma(\cdot)$ is a sigmod activation function.
The candidate state $\widetilde{h}_{ij}$ is computed in a way similar to a traditional RNN:
\begin{small}
	\begin{gather*}
	\widetilde{h}_{ij}=\mathrm{tanh}(W_{h}x_{ij}+reset_{ij}\odot (U_{h}h_{i,j-1})+b_{h}),\tag{3}
	\end{gather*}
\end{small}
where $\mathrm{tanh}(\cdot)$ is a tanh activation function.

Reset gate $reset_{ij}$ controls how much the past state contributes to the candidate state $\widetilde{h}_{ij}$. Its update function is as follows:
\begin{small}
	\begin{gather*}
		reset_{ij}=\sigma(W_{reset}x_{ij}+U_{reset}h_{i,j-1}+b_{reset}).\tag{4}
	\end{gather*}
\end{small}

Different relations may contribute differently to the semantic representation of the multi-hop path.
Hence, we introduce the attention mechanism~\cite{Bahdanau2014axriv} to extract the important relations and aggregate the representation of those meaningful relations to form a specific output.
As illustrated in the relation attention layer in Figure~\ref{fig:overall}, we utilize a single attention layer to get $u_{ij}$ as a hidden representation of $h_{ij}$:
\begin{small}
	\begin{gather*}
	{u_{ij}}=\mathrm{tanh}(W_{e}h_{ij}+b_{e}).\tag{5}
	\end{gather*}
\end{small}
Specifically, we measure the importance of the relation as the similarity/relatedness of $u_{ij}$ with a context entity vector $u_{e}$, and get a normalized importance weight $a_{ij}$ using a softmax function as
\begin{small}
	\begin{gather*}
	a_{ij}=\frac{\mathrm{exp}(u^{\top}_{ij}u_{e})}{\sum^{T_{i}}_{k=1}\mathrm{exp}(u^{\top}_{ik}u_{e})},\tag{6}
	\end{gather*}
\end{small}
where $\mathrm{exp}(\cdot)$ is an exponential function based on the natural constant $e$.
Then, we obtain the representation of the multi-hop path by weighted summing of the relation annotations as follows:
\begin{small}
	\begin{gather*}
	p_{i}=\sum^{T_{i}}_{j=1}a_{ij}h_{ij}.\tag{7}
	\end{gather*}
\end{small}
As shown in the path encoder in Figure~\ref{fig:overall}, to mine the shared semantic features between single relation and multi-hop paths, we use a weighted sum over all the encoded path vectors between entities.
As the process and calculation methods are similar to the relation encoder, we omit the detailed description for simplicity.
So far, we selectively encode all multi-hop paths in $\mathcal{P}$ into a low-dimensional vector based on HANs.

\subsubsection{Coupled Neural Networks}\label{cnn}
In the proposed method, we define two feed-forward architectures to discriminate the source of the extracted feature and its corresponding relation.
As shown in the right part of Figure~\ref{fig:overall}, the two neural networks are coupled together by a sub-module that extracts features from different input sources.
Under the assumption mentioned above, these extracted features are considered as the semantic information shared by the encoded multi-hop paths $p$ and the encoded single relation $r$.
We decompose such coupled neural networks into three parts to detail each sub-module.

\textit{\textbf{\romannumeral1. Parameter-Shared Feature Extractor:}}
As a sub-module shared by two neural networks, the feature $f$ extracted by the feature extractor should minimize the classification error and be source-indiscernible in the meantime.
In the proposed model, the feature mapping is accomplished by several feed-forward layers, which are similar to the encoder in stacked AutoEncoder~\cite{autoencoder2010jmlr}.
Because this sub-module is shared by the follow-up discriminator and classifier, all of the parameters $\theta_{e}$ in the feature extractor $\mathcal{E}(x;\theta_{e})$ are updated based on the gradients from downstream parts (as illustrated in the upper $\mathcal{L}^{C}$ and bottom $\mathcal{L}^{D}$ in Figure~\ref{fig:overall}).
To improve the ability of the feature extractor to resist noise and remain sparse, we add sparse constraints in the loss function.
After which, the loss function of the feature extractor is defined as
\begin{small}
	\begin{gather*}
		\mathcal{L}^{E}=\mathcal{L} + \beta \sum_{j}KL(\rho||\hat{\rho}_{j}),\tag{8}
	\end{gather*}
\end{small}
where $\mathcal{L}$ is the downstream loss (i.e. the combination of $\mathcal{L}^{C}$ and $\mathcal{L}^{D}$), $\beta$ denotes the constraint rate, $KL(\cdot)$ is the KL divergence, and $\rho$ and $\rho_{j}$ represent the expected activation of neurons and average activation degree on the training set.

In general, for the source discriminator, the feature extractor can be considered as a query of `what are the source-shared features?'
Thus from the classifier's perspective, it can be regarded as a query of `what are the valuable features for determining relation?'
Furthermore, it also stands for a high-level representation of a fixed query in the whole model of `what are the pivots between a source discriminator and relation classifier?'
More theoretically, to ensure that the features extracted by the feature extractor can be obtained from the single relation and the multi-hop paths, and can effectively correspond to the true single relation, this paper selects a coupled adversarial structure to achieve the above two constraints.
Among them, the source discriminator confuses features from different sources by providing a reversal gradient, so that the features obtained by the feature extractor from different sources obey a similar distribution.
The relation classifier can still accurately correspond to the unique correct relation by the similar features obtained from the single relation and the multi-hop path through the classification error constraint.

\textit{\textbf{\romannumeral2. Source Discriminator:}}
The source discriminator is a sub-module exploited to discriminate the source of input features.
Its discriminant results can be used to optimize the performance of the feature extractor.	
As illustrated in the bottom right part of Figure~\ref{fig:overall}, we treat the shared features $f \in \{f_{r},f_{p}\}$ as a credential that determines the source of the input features which is computed as
\begin{small}
	\begin{gather*}
		z_{i}=W_{di}f+b_{di}, y_{i}=\frac{\mathrm{exp}(z_{i})}{\sum^{2}_{j=1}\mathrm{exp}(z_{j})},\tag{9}
	\end{gather*}
\end{small}
where $z_{i},i \in [0,1]$ is the independent probability of each data source, $y_{i}$ is the normalized probability for different data sources.
The goal of the source discriminator is to minimize the cross-entropy loss within all data distribution:
\begin{small}
	\begin{gather*}
		\mathcal{L}^{D}=-\frac{1}{2N_{s}}\sum^{2N_{s}}_{k=1}(\hat{y_{k}}\mathrm{ln}y_{k}+(1-\hat{y_{k}})\mathrm{ln}(1-y_{k})),\tag{10}
	\end{gather*}
\end{small}
where $\hat{y_{k}}\in \{0,1\}$ denotes the real source of sample $k$ and $N_{s}$ is the number of positive samples.

To ensure that the features extracted by the feature extractor are source-indiscernible, we introduce \textit{Gradient Reversal Layer} (GRL)~\cite{ganin2015icml} between the feature extractor and the source discriminator, as shown in the bottom right part of Figure~\ref{fig:overall}.
The GRL achieves gradient reversal through multiplying the gradient by a certain negative constant during the back propagation-based training, which guarantees similar feature distributions over the two sources.
Specifically, GRL acts as an identity transform during the forward propagation, but takes the gradient from subsequent level, multiplies it by $-\lambda$, and passes it to the preceding layer.
In our framework, we treat the GRL as a “pseudo-function” $R_{\lambda}(x)$ defined by the following equations, describing its forward- and back-propagation behaviors:
\begin{small}
	\begin{gather*}
	R_{\lambda}(x)=x, \frac{\partial R_{\lambda}(x)}{\partial x}=-\lambda I,\tag{11}
	\end{gather*}    
\end{small}
where $I$ is an identity matrix and $\lambda$ is the adaptation rate.
Hence, the independent possibility $z_{i}$ has to be rewritten as 
\begin{small}
	\begin{gather*}
	z_{i}=W_{di}R_{\lambda}(f)+b_{di}.\tag{12}
	\end{gather*}
\end{small}
Mathematically, we treat the source discriminator as $\mathcal{D}(R_{\lambda}(f);\theta_{d})$ and the feature extractor as $\mathcal{E}(x;\theta_{e})$, where $\theta_{d}$ and $\theta_{e}$ are corresponding parameters.    
As illustrated in Figure~\ref{fig:overall}, the $\theta_{d}$ is optimized to enhance the capability of the source discriminator to distinguish different data sources, while $\theta_{e}$ is trained to fool the discriminator based on the reversal gradient $-\lambda \frac{\partial \mathcal{L}^{D}}{\partial \theta_{e}}$.

\textit{\textbf{\romannumeral3. Relation Classifier:}}
This sub-module is used to determine the missing relation between the given entity pairs.
It outputs the probability of each candidate relation in relation set $R$.
As illustrated in the top right part of Figure~\ref{fig:overall}, we treat the output from the feature extractor as the representative features to determine the single relation and feed it to the softmax layer for relation classification.
Because the calculation method of the softmax layer is similar to that of the corresponding part in the source discriminator, we omit it.
The loss function for the relation classifier can be written as
\begin{small}
	\begin{gather*}
	\mathcal{L}^{C}=-\frac{1}{N_{s}}\sum^{N_{s}}_{k=1}\sum^{N_{c}}_{i=1}(\hat{y_{i}}\mathrm{ln}y_{k\_i}),\tag{13}
	\end{gather*}
\end{small}
where $\hat{y_{i}}\in \{0,1\}$ and $y_{k\_i}$ are the ground truth and output result respectively.

\subsubsection{Regularization}
In order to avoid over-fitting, the squared Frobenius norm and squared $\ell_{2}$ norm for different weight $W_{d},W_{c}$ and bias $b_{d},b_{c}$ are added as 
\begin{small}
	\begin{gather*}
	\mathcal{L}^{R}=||W_{d}||^{2}_{F}+||b_{d}||^{2}_{2}+||W_{p}||^{2}_{F}+||b_{p}||^{2}_{2},\tag{14}
	\end{gather*}
\end{small} 
where $||\cdot||_{2}$ and $||\cdot||_{F}$ respectively denote the $\ell_{2}$ norm of a vector and Frobenius norm of a matrix.

\subsection{Joint Adversarial Training}

The overall objective function of the proposed model is a combination of the losses of its components.
\begin{small}
	\begin{gather*}
	\mathcal{L}^{Total}=\mathcal{L}^{D}+\mathcal{L}^{C}+\beta \sum_{j}KL(\rho||\hat{\rho}_{j})+\rho_{r} \mathcal{L}^{R},\tag{15}
	\end{gather*}
\end{small} 
where $\rho_{r}$ is the regularization parameter.
To minimize $\mathcal{L}^{Total}$, we use a joint AT strategy to iteratively update all sub-modules.

Before AT, we pre-train the components with specific methods to avoid possible problems (e.g., gradient vanishing, and gradient instability)~\cite{arjovsky2017towards}.
The implementation details (including initialization, determining parameters, and pre-training) of each component will be detailed in the follow-up section.

During AT, the source discriminator\footnote{To ensure the performance of the source discriminator, we manually adjust the input distribution in each mini-batch to balance the number of inputs from the two sources.} and the relation classifier can be directly trained to minimize their loss by using stochastic gradient descent (SGD) based method as
\begin{small}
	\begin{gather*}
	\theta_{c} := \theta_{c} + \mu \frac{\partial \mathcal{L}^{C}}{\partial \theta_{c}},
	\theta_{d} := \theta_{d} + \mu \frac{\partial \mathcal{L}^{D}}{\partial \theta_{d}},\tag{16}
	\end{gather*}
\end{small}
where $\theta_{c}$ and $\theta_{d}$ denote the parameters in the classifier and discriminator respectively, and $\mu$ is the learning rate.
As the upstream module, the feature extractor has multiple gradient sources.
Hence, as illustrated in the bottom part of Figure~\ref{fig:overall}, reversed gradient (through passing the GRL) from the discriminator and back-propagation gradient from the classifier are used to update the parameters by fine-tuning as
\begin{small}
	\begin{gather*}
	\theta_{e} := \theta_{e} + \mu \frac{\partial \mathcal{L}^{E}}{\partial \theta_{e}}.\tag{17}
	\end{gather*}
\end{small}
In particular, in order to ensure the performance of AT, we balance the distribution of gradient source in each batch.

\subsection{Analysis of Model Perspectives}\label{per}
From the perspective of the whole model, the motivation of our model is to ensure that the shared features extracted from different sources are valuable for completing missing relations.
In the following, we propose some alternative perspectives to facilitate understanding of our model.

\textbf{Information Theory:}
Consider $X_{r} \sim q(X_{r})$ is the distribution of $r$, $q$ unknown, $X_{f_{r}}=f_{\theta_{e}}(X_{r})$. 
It can easily be shown that minimizing the expected classification error and discrimination error amounts to respectively maximize a lower bound on mutual information $I(X_{f_{r}};X_{r})$ and $I(X_{f_{r}};X_{f_{p}})$.
Because downstream tasks are independent of each other, our model can thus be justified by the objective that $f_{\theta_{e}}(\cdot)$ captures as much source-indiscernible information as possible for relation classification, i.e. maximizing a lower bound on $I(X_{r}; X_{f_{p}})$.

\textbf{Manifold Learning:}
We treat $f$ as a low-dimensional manifold, and $f_{\theta_{e}}(\cdot)$ is a stochastic operator learnt to map $r,p$ to $f_{r},f_{p}$ which tends to go from lower probability points to high probability points, generally on or near the manifold.
Our model can thus be seen as a way to define and learn a manifold, and the intermediate representation $Y=f_{\theta_{e}}(X)$ can be interpreted as a coordinate system for points on the manifold.
More generally, one can think of $Y=f_{\theta_{e}}(X)$ as a representation of $X$ which is well suited to capture the \textbf{valuable} (not \textbf{main}) variations in the input, i.e. on the manifold. 
When additional criteria (i.e. source indiscrimination) are introduced in the learning model, one can no longer directly view $Y=f_{\theta_{e}}(X)$ as an explicit low-dimensional coordinate system for points on the manifold, but it retains the property of capturing the specific factors of source-shared variation in the data.

\textbf{Generative Model:}
As our model takes an input vector $x \in \{r\} \bigcup \{p\}, x \in \mathbb{R}^{d}$, and maps it to features $f\in\{f_{r}\}\bigcup\{f_{p}\}, f\in \mathbb{R}^{d'}$ through feature extractor $f=f_{\theta_{e}}(x)$, parameterized by $\theta_{e}$.
Depending on the different downstream tasks, feature $f$ is then mapped back to the `reconstructed' outputs, i.e. relation $r$ (through relation classifier) and source (through source discriminator) by $y_{c}=g_{\theta_{c}}(f_{r})$ and $y_{d}=h_{\theta_{d}}(f)$.
Each training $x^{(i)}$ is mapped to a corresponding $f^{(i)}$, an output relation $y_{c}$ (only for $x_{r}$) and a source $y_{d}$.
Since the downstream tasks are pre-trained, the parameters of the model are optimized to minimize the \textit{average reconstruction error} and maximize the \textit{source discrimination error}:
\begin{small}
	\begin{gather}
	\theta_{e}^{\star}=\mathop{\arg\min}_{\theta_{e}}(\frac{1}{m}\sum_{i=1}^{m}L(x^{(i)},y_{c}^{(i)})-\frac{1}{n}\sum_{j=1}^{n}L(x^{(j)},y_{d}^{(j)})),\tag{18}
	\end{gather}
\end{small}
where $L$ is a loss function, $m=|\{r\}|, n = |\{r\}\bigcup\{p\}|$.

As mentioned in Section~\ref{cnn}, $L$ is the \textit{cross-entropy}. i.e. $L=L_{\mathbb{H}}$, then:
\begin{small}
	\begin{gather}
	\theta_{e}^{\star}=\mathop{\arg\min}_{\theta_{e}}\mathbb{E}_{q^{0}(X)}(L_{\mathbb{H}}(X,Y_{c})-L_{\mathbb{H}}(X,Y_{d})),\tag{19}
	\end{gather}
\end{small}
where $q^{0}(X)$ denotes the empirical distribution associated to training inputs, $\mathbb{E}_{p(X)}[f(X)]=\int p(x)f(x)\mathrm{dx}$ is the expectation of $f(X)$.

Specifically, we recover the training criterion for our model from a generative model perspective. Then the goal of our model is to learn a constrained joint probability distribution, i.e. the co-occurrence probability of the input data $X$, the source-indiscernible feature $f$, and the relation $r$.
Consider the generative model $p(X,f,Y_{c},\bar{Y_{d}})=p(Y_{c},\bar{Y_{d}})p(X|Y_{c},\bar{Y_{d}})p(f|X))$, where $\bar{Y_{d}}$ denotes the wrong source label. 
We can show that training the feature extractor as described above is equivalent to maximizing a variational bound on that generative model:
\begin{small}
	\begin{align}
	&\mathop{\arg\max}_{\theta_{e}}\mathbb{E}_{q^{0}(X)}[\log p(X,f,Y_{c},\bar{Y_{d}})]\notag\\
	=&\mathop{\arg\max}_{\theta_{e}}\mathbb{E}_{q^{0}(X)}[\log p(Y_{c},\bar{Y_{d}})p(X|Y_{c},\bar{Y_{d}})p(f|X))].\tag{20}
	\end{align}
\end{small}
$p(Y_{c},\bar{Y_{d}})$ is a uniform prior probability and $q^{0}(X|f) \propto q^{0}(f|X)q^{0}(X)$, then
\begin{small}
	\begin{align}
	&\mathop{\arg\max}_{\theta_{e}}\mathbb{E}_{q^{0}(X)}[\log p(Y_{c},\bar{Y_{d}})p(X|Y_{c},\bar{Y_{d}})p(f|X))]\notag\\
	=&\mathop{\arg\max}_{\theta_{e}}\mathbb{E}_{q^{0}(X)}[\log p(X|Y_{c},\bar{Y_{d}})]\notag\\
	=&\mathop{\arg\max}_{\theta_{e}}\mathbb{E}_{q^{0}(X)}[\log (p(X|Y_{c}=g_{\theta_{c}}(f_{r})) \cdot p(X|\bar{Y_{d}}=h_{\theta_{d}}(f)))]\notag\\
	=&\mathop{\arg\min}_{\theta_{e}}\mathbb{E}_{q^{0}(X)}(L_{\mathbb{H}}(X,Y_{c})-L_{\mathbb{H}}(X,Y_{d})).\tag{21}
	\end{align}
\end{small}
Obviously, this is consistent with our training goals, i.e. Equation 19.

\section{Experiments}\label{s4}
We evaluate the proposed method empirically in terms of classification effect and model interpretability.

\begin{table*}[t]
	\small
	\centering
	\caption{Statistics of the five datasets. Since we only retain entity pairs that contain multi-hop paths, there are fewer entities than in the original datasets.}
	\label{tab:dataset}
	\begin{tabular}{lccccc}		
		\hline
		& FB15K* & FB15K-237* & FBe30K* & WN18* & WN18RR* \\		
		\hline
		\#Entities & 13126 & 10463 & 19835 & 21569 & 21562 \\
		\#Entity Pairs & 37785 & 18473 & 64732 & 77396 & 77092 \\
		\#Relations & 1345 & 237 & 1493 & 18 & 11 \\
		Average Multi-hop Paths ($length \leq3$) between Each Entity Pair & 11.62 & 9.34 & 12.85 & 8.63 & 8.59 \\
		\hline
	\end{tabular}
\end{table*}

\subsection{Datasets}
Since our method requires an abundant amount of multi-hop paths between entities, we use standardized FB15K and WN18 as experimental datasets~\cite{Bordes2013nips}.
Among them, FB15K is a highly dense sub-graph captured from Freebase, which contains triples composed by two entities and a relation.
WN18 is a set of linguistic triples obtained from WordNet, which is a lexical English dictionary containing linguistic relation between words, e.g., \texttt{hypernym, hyponym} and \texttt{meronym}.
We also set up a delicate sub-graph with some high-quality, intuitively inferable sub-graphs\footnote{The sub-graphs mainly include triples whose relations can be intuitively infer ed, e.g., \texttt{Children}, \texttt{Spouse}.} from FreebaseEasy~\cite{bast2014www} as the third experimental dataset (namely FBe30K), to further validate the hierarchical attention mechanism in a relatively controllable context.
Furthermore, we also use FB15K-237 and WN18RR as experimental data, which is a filtered version of FB15K and WN18~\cite{fb15k-2372015-,conv2d2018aaai} where reciprocal triples are removed.
To train the model, we select the entity pairs with single relation in between and having connecting multi-hops paths, and divide them into training, validation, and test sets by 8:1:1.
Note that we add annotation * to emphasize that our datasets have been restructured.
Table~\ref{tab:dataset} details the statistical information of the datasets.

\subsection{Experimental Setup}
We evaluate the the performance of the models by two widely-used link prediction testing protocols: the Mean Rank (MR) of the original triple among corrupted ones, and the proportions of valid entities ranked in top 1\%, 3\%, and 10\% (Hit@1, Hit@3, Hit@10).
Following \cite{Bordes2013nips}, we also report \textit{filtered} results to avoid underestimations caused by corrupted triples that are in fact valid.

\begin{table*}[ht]
	\scriptsize
	\centering
	\caption{Comparison of different methods for relation classification in FB15K*, FB15K-237*, and FBe30K*. Higher Hits@n and lower Mean Rank indicate better predictive performance.}
	\label{tab:result}
	\begin{tabular}{llcccccccccccc}
		\hline
		& \multirow{2}{*}{Model} & \multicolumn{4}{c}{FB15K*} & \multicolumn{4}{c}{FB15K-237*} & \multicolumn{4}{c}{FBe30K*} \\
		& & MR(\textit{filter}) & Hits@1 & Hits@3 & Hits@10 & MR(\textit{filter}) & Hits@1 & Hits@3 & Hits@10 & MR(\textit{filter}) & Hits@1 & Hits@3 & Hits@10 \\
		\hline
		\multirow{7}{*}{\rotatebox{90}{\textbf{T}}} & TransE & 2.71 & 89.14 & 90.70 & 91.27 & 3.54 & 83.88 & 85.96 & 87.17 & 2.67 & 89.63 & 90.83 & 91.51 \\
		& TransH & 2.50 & 90.06 & 91.63 & 92.43 & 3.34 & 84.64 & 87.08 & 88.21 & 2.48 & 90.37 & 91.87 & 92.52 \\
		& TransR & 2.33 & 91.25 & 92.75 & 93.27 & 3.19 & 85.42 & 87.71 & 89.08 & 2.32 & 91.58 & 92.67 & 93.34 \\
		& CTransR & 2.10 & 92.30 & 93.86 & 94.54 & 3.11 & 86.09 & 88.18 & 89.43 & 2.09 & 92.56 & 93.92 & 94.51 \\
		& DistMult & 1.61 & 94.71 & 96.43 & 97.19 & 2.63 & 88.47 & 90.90 & 91.98 & 1.57 & 94.91 & 96.72 & 97.31 \\
		& ComplEx & 1.59 & 94.82 & 96.63 & 97.23 & 2.57 & 88.49 & 90.97 & 92.34 & 1.56 & 95.73 & 96.75 & 97.33 \\
		& ConvE & 1.46 & 95.63 & 97.34 & 97.84 & 2.40 & 89.40 & 91.84 & 93.26 & 1.45 & 95.84 & 97.30 & 97.98 \\
		& RotatE & 1.44 & 95.67 & 97.75 & 97.98 & 2.36 & 89.77 & 92.01 & 93.25 & 1.36 & \textbf{96.83} & 97.64 & 98.42 \\
		\hline
		\hline
		\multirow{4}{*}{\rotatebox{90}{\textbf{T\&P}}}& rTransE & 1.87 & 93.51 & 95.17 & 95.71 & 2.77 & 87.69 & 90.10 & 91.26 & 1.86 & 94.06 & 95.21 & 95.73 \\
		& PTransE(MUL 2-step) & 1.93 & 93.04 & 94.88 & 95.42 & 2.95 & 86.98 & 88.99 & 90.30 & 1.92 & 93.38 & 94.86 & 95.46 \\
		& PTransE(RNN 2-step) & 1.58 & 94.74 & 96.64 & 97.29 & 2.52 & 88.98 & 91.24 & 92.62 & 1.57 & 95.21 & 96.71 & 97.28 \\
		& PTransE(ADD 2-step) & 1.50 & 95.27 & 97.08 & 97.65 & 2.43 & 89.56 & 91.78 & 93.05 & 1.50 & 95.85 & 97.11 & 97.68 \\
		\hline
		\hline
		\multirow{6}{*}{\rotatebox{90}{\textbf{P}}} & B-PR & 2.52 & 89.72 & 91.62 & 92.35 & 3.46 & 84.36 & 86.48 & 87.54 & 2.51 & 90.49 & 91.73 & 92.38 \\
		& SFE-PR & 2.34 & 90.87 & 92.69 & 93.24 & 3.25 & 85.05 & 87.49 & 88.75 & 2.32 & 91.32 & 92.67 & 93.36 \\
		& Ours (Word2Vec\&Shortest) & 2.13 & 91.82 & 93.75 & 94.40 & 2.86 & 87.37 & 89.67 & 90.73 & 2.09 & 92.52 & 93.93 & 94.52 \\
		& Ours (Word2Vec\&RW) & 1.98 & 92.78 & 94.45 & 95.14 & 2.74 & 87.78 & 90.26 & 91.38 & 1.94 & 93.38 & 94.74 & 95.35 \\	
		& Ours (TransE\&Shortest) & 1.50 & 95.10 & 97.13 & 97.68 & 2.49 & 88.98 & 91.32 & 92.80 & 1.43 & 96.05 & 97.39 & 98.07 \\
		& Ours (TransE\&RW) & \textbf{1.37} & \textbf{95.81} & \textbf{97.78} & \textbf{98.37} & \textbf{2.32} & \textbf{89.89} & \textbf{92.28} & \textbf{93.65} & \textbf{1.33} & 96.55 & \textbf{97.97} & \textbf{98.50} \\
		\hline
	\end{tabular}
\end{table*}

\subsubsection{Baseline Methods}
We select three groups of models and compare them in the reconstructed datasets\footnote{We obtain the experimental results of all baseline methods by rerunning their codes.}.

\textit{Triple information based methods} (Group \textbf{T})\footnote{We only use the selected entity pair (i.e. triple) for training, so there is a gap between our results and those in the original papers.}: 
TransE models relations between entities by interpreting them as translations operating on the low-dimensional embeddings of the entities~\cite{Bordes2013nips}.
TransH models a relation as a hyperplane together with a translation operation on it and proposes an efficient method to construct negative examples to reduce false negative labels in training~\cite{Wang2014aaai}.
TransR builds entity and relation embeddings in separate entity and relation spaces and learns embeddings by first projecting entities from entity space to corresponding relation space, and then building translations between projected entities~\cite{Lin2015aaai}.
CTransR is cluster-based TransR, which clusters diverse head-tail entity pairs into groups and learns distinct relation vectors for each group~\cite{Lin2015aaai}.

DistMult implements triple modeling by defining triple as a matrix operation~\cite{distmult2015iclr}.
ComplEx introduces complex space to optimize single relationship modeling~\cite{complex2016icml}.
ConvE uses 2D convolution to model the relationship between entity pairs~\cite{conv2d2018aaai}.
RotatE learns knowledge embedding through relational rotation in complex space~\cite{rotate2019axriv}.

\textit{Triple and Path information based methods} (Group \textbf{T\&P}): 
rTransE learns to explicitly model the composition of relationships via the addition of their corresponding translation vectors~\cite{Garcdur2015emnlp}.
PTransE considers relation paths as translations between entities for representation learning and addresses problems of path reliability and semantic composition~\cite{lin2015emnlp}.
Different methods in multi-hop paths modeling can be divided into (RNN 2-step), (ADD 2-step), and (MUL 2-step).

\textit{Path information based methods} (Group \textbf{P}): 
B-PR considers binarized path features for learning PR-classifier~\cite{Gardner2015emnlp}.
SFE-PR is a PR-based method that uses breadth-first search with Subgraph Feature Extraction (SFE) to extract path features~\cite{Gardner2015emnlp}. In particular, SFE-PR uses only PR-style features for feature matrix construction.
Ours is the model proposed in this paper.

\subsubsection{Implementation Details}
As we found that various dimensions of relation embeddings and PE resulted in similar trends, only experimental results for 100 dimension relation embeddings and 5 dimension PE will be reported here.
To construct path set $\mathcal{P}$, we utilize the shortest path principle $min\{3, n\}$ and RW strategy~\cite{Lao2015acl} respectively.
We initialize the input embeddings with two different pre-trained embeddings (i.e. Word2Vec~\cite{mikolov2013distributed} and TransE~\cite{Bordes2013nips}), and randomly initialize other model parameters, e.g., attention vectors, feature extractor and softmax layers of relation classifiers and source discriminators.

Each module is pre-trained after initialization, in which the training target of the whole model (except source discriminator) is to minimize the classification error of the relation classifier.
The source discriminator is trained from scratch after the classifier converges.
The hyperparameters of the model are chosen by maximizing Mean Rank on the validation set: the regularization weight $\rho_{r}$ is set to 0.05, the constrain rate $\beta$ is set to 0.01 and the expected activation degree $\rho$ is set to 0.05.
The adaptation factor $\lambda$ in GRL is gradually changed from 0 to 1 during the training process as $\lambda = \frac{2}{1+\mathrm{exp}(\frac{\gamma \cdot n_{c}}{nN_{s}})}-1$~\cite{ganin2015icml}.
Our model is optimized with the SGD over shuffled mini-batches with momentum rate 0.95, learning rate is decayed as $\eta = \frac{0.005}{(1+\frac{\gamma \cdot n_{c}}{nN_{s}})^{0.5}}$, where $n_{c}$ and $n$ respectively denote the number of trained samples and iterations, and $\gamma$ is set to 10.
The batch sizes for the source discriminator and relation classification are both 100, and the batch for source discriminator coming from different sources equally.

\subsection{Experimental Result and Analysis}
We compare our method with baselines in two aspects as discussed below.

\subsubsection{Classification Effect}
Table~\ref{tab:result} and Table~\ref{tab:result2} show the comparative results of classification effect on FB and WN respectively.
We can observe that our method outperforms the other two groups of baseline methods significantly, except that Hits@1 is slightly lower than RotatE.
With the use of different pre-trained relation embeddings and a more reasonable path selection algorithm, the performance of our model is enhanced greatly.
\begin{figure*}[t]
	\centering
	\includegraphics[width=1\textwidth]{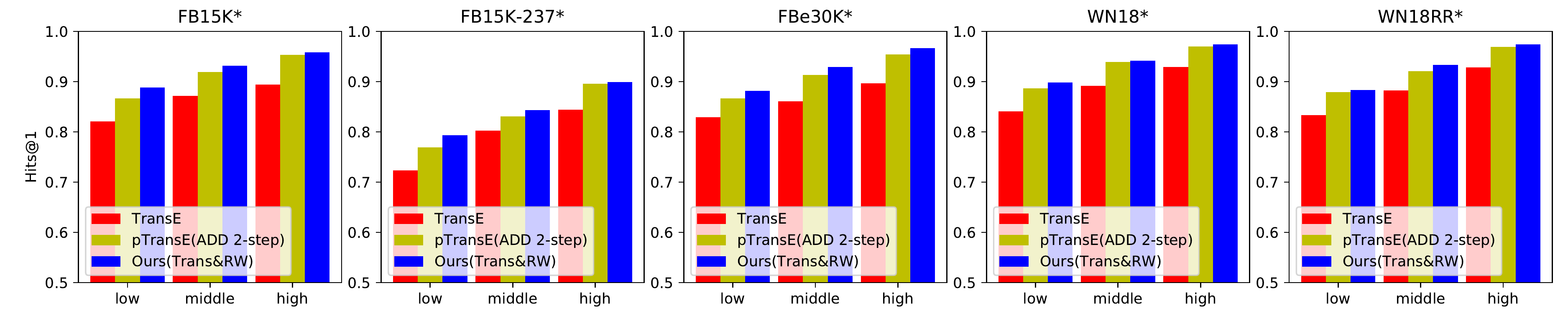}
	\caption{
		The relation completion results of the three methods in different frequency relations.
		We chose PTransE (which also introduced path information) and TransE (which is our initial inputs) to compare our method.
		(low, middle, high) means the distribution of relation frequency.
	}
	\label{fig:bucket}
\end{figure*}

\begin{figure*}[t]
	\begin{minipage}[t]{0.33\linewidth}
		\centering
		\includegraphics[width=1\textwidth]{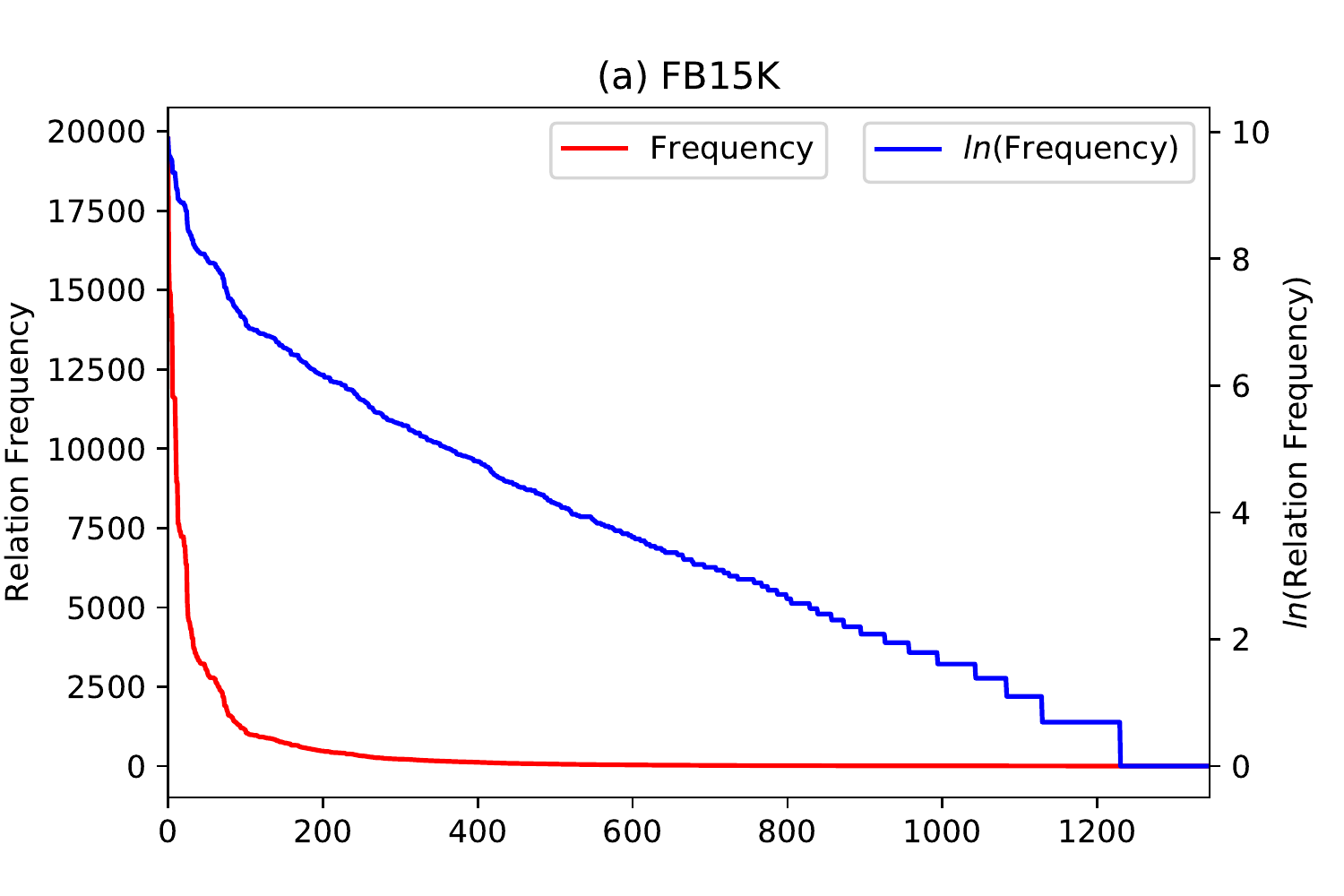}
		\centerline{(a) \small{FB15K}}
	\end{minipage}
	\hspace{0.005\linewidth}
	\begin{minipage}[t]{0.33\linewidth}
		\centering
		\includegraphics[width=1\textwidth]{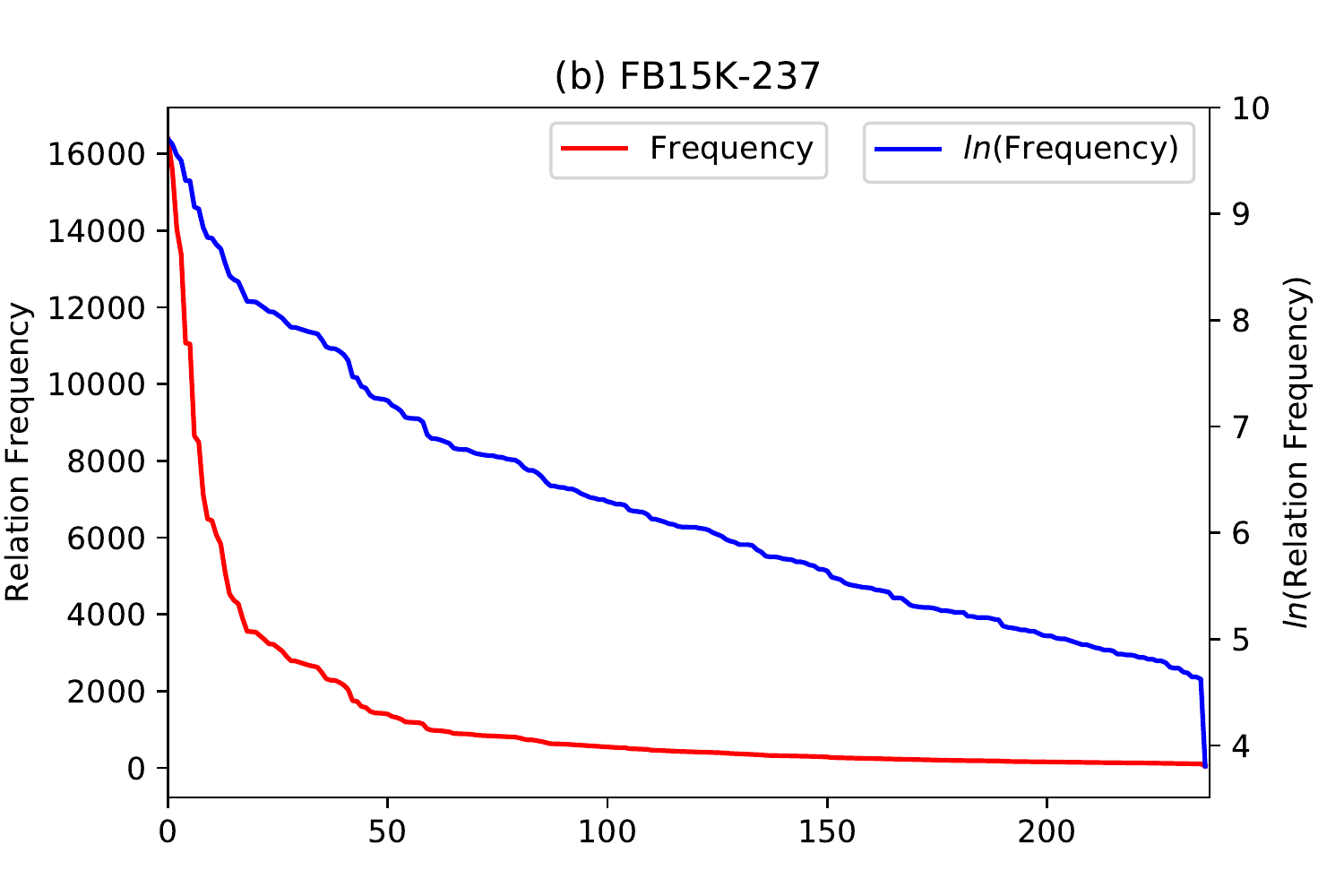}
		\centerline{(b) \small{FB15K-237}}
	\end{minipage}
	\hspace{0.005\linewidth}
	\begin{minipage}[t]{0.33\linewidth}
		\centering
		\includegraphics[width=1\textwidth]{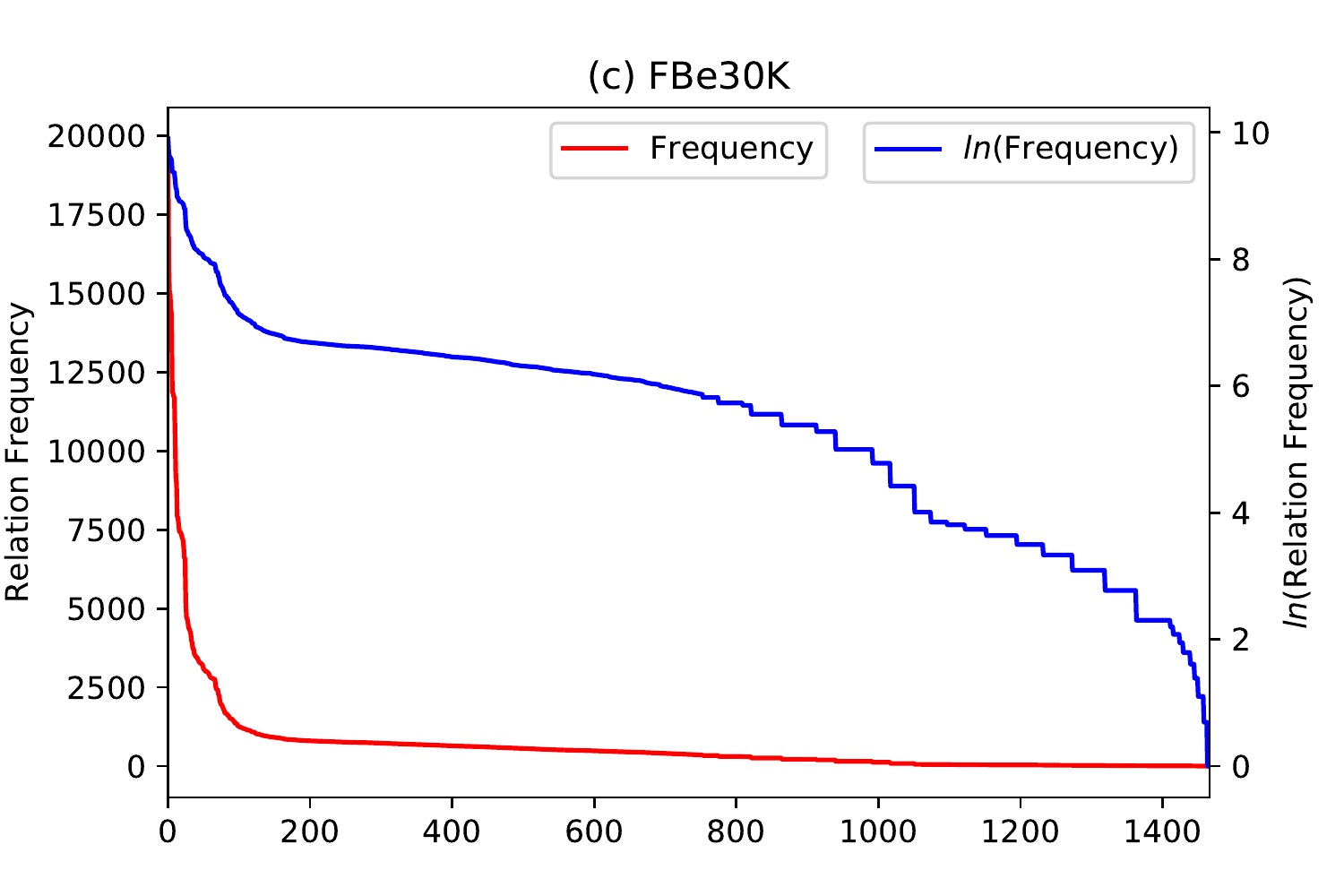}
		\centerline{(c) \small{FBe30K}}
	\end{minipage}
	\hspace{0.005\linewidth}
	\begin{minipage}[t]{0.33\linewidth}
		\centering
		\includegraphics[width=1\textwidth]{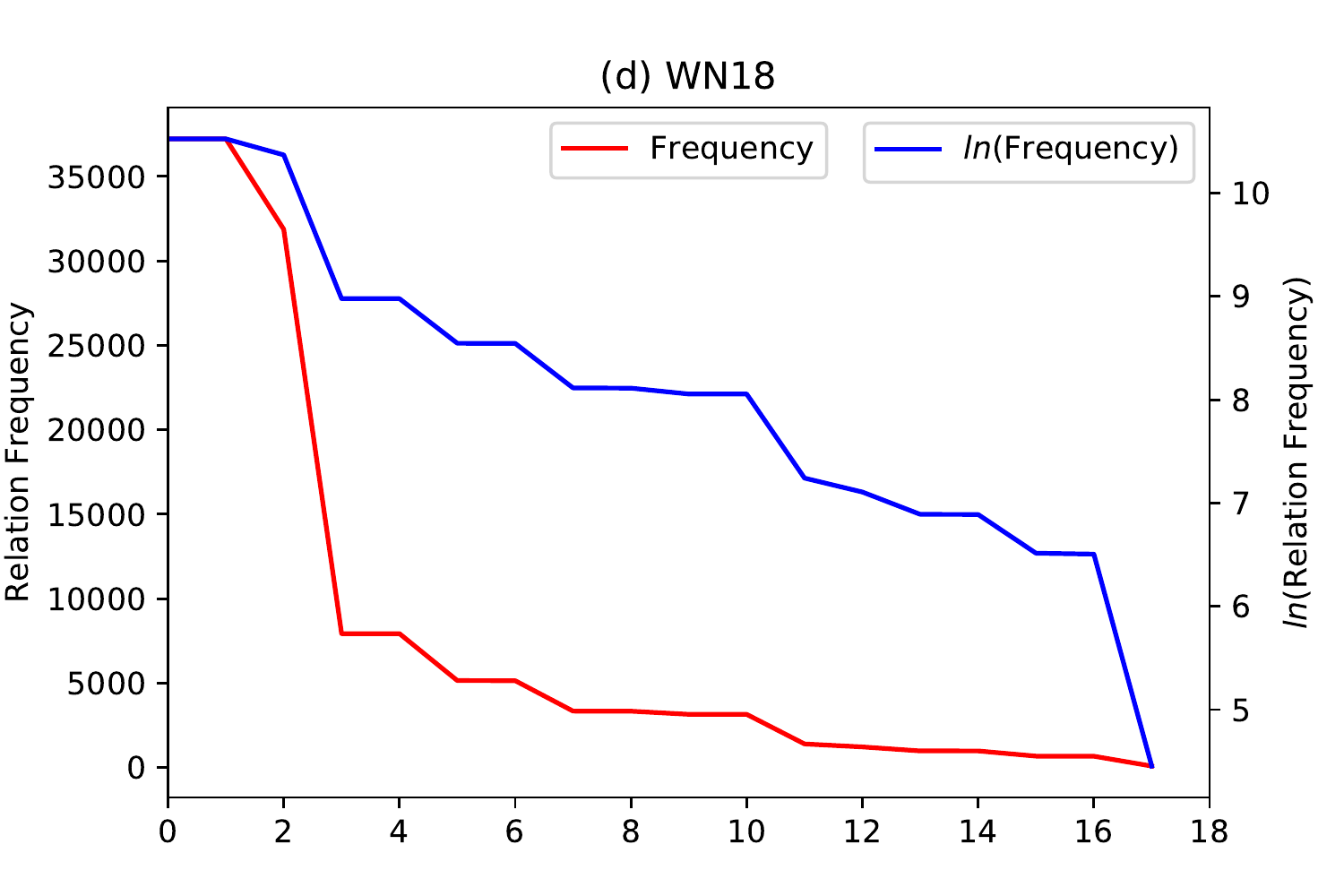}
		\centerline{(d) \small{WN18}}
	\end{minipage}
	\hspace{0.005\linewidth}
	\begin{minipage}[t]{0.33\linewidth}
		\centering
		\includegraphics[width=1\textwidth]{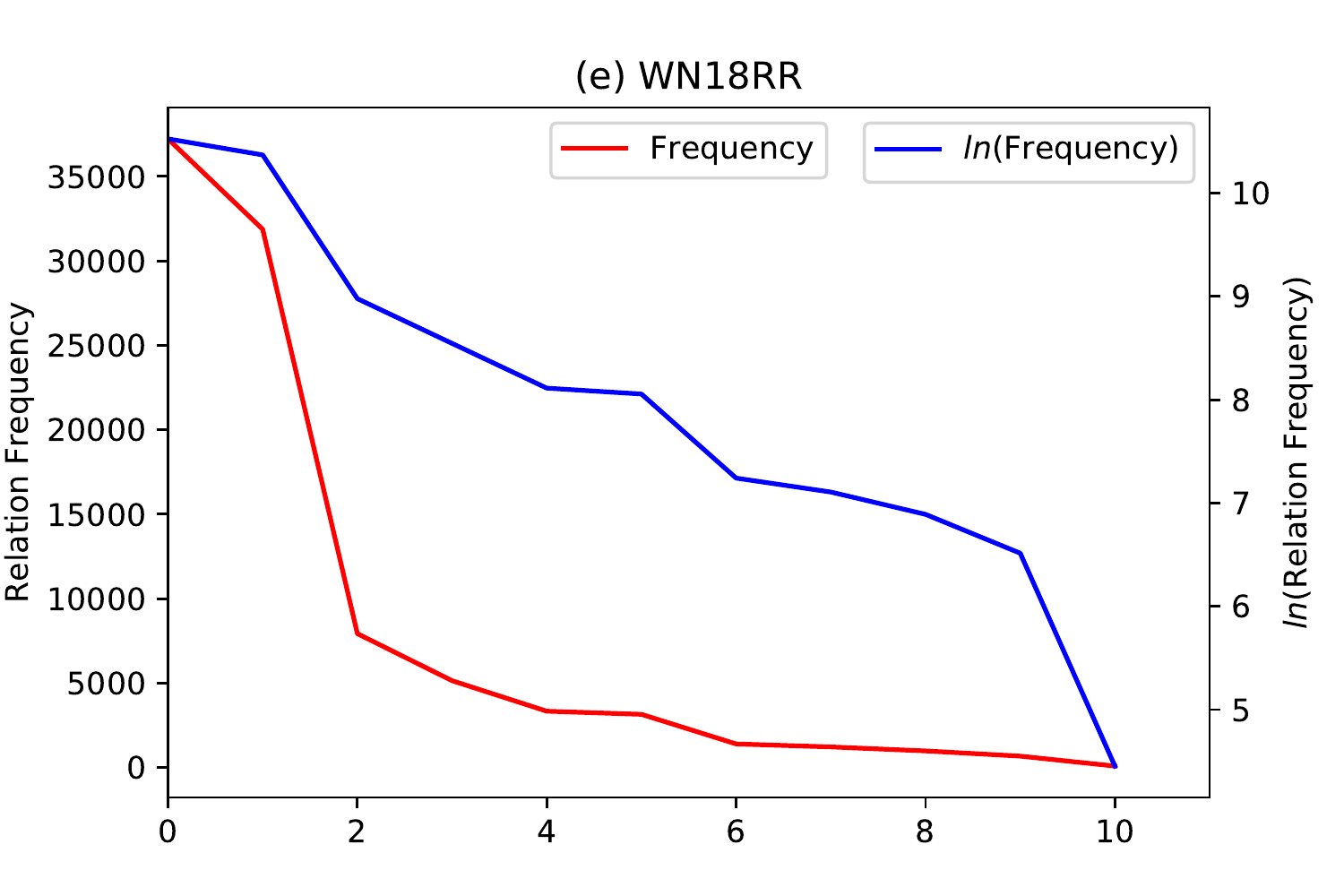}
		\centerline{(e) \small{WN18RR}}
	\end{minipage}
	\caption{
		Frequencies and log frequencies of single relations in three datasets.
		The x-axis is single relations sorted by frequency. 
	}
	\label{fig:distribution}
\end{figure*}

In contrast to the methods in Group \textbf{T}, we find that our method (TransE\&RW) improves the classification effect in both MR and Hits@ 3, 10. 
Furthermore, despite the fact that rTransE and PTransE in Group \textbf{T\&P} introduce path information into original TransE by using additional translation vectors, the overall effect is still not as good as our model's. 
This indicates that although the introduction of path information can significantly improve the classification effect, its stage and strategy can affect the final results.

By observing the performances on different datasets, we can find that the prediction effects of all models (especially B-PR and SFE-PR) become worse after removing reverse relations.
However, since our model can handle multi-hop paths efficiently, the loss of a one-hop reverse path has less impact on the final results.

\begin{table}[t]
	\scriptsize
	\centering
	\caption{Comparison of different methods for relation classification in WN18* and WN18RR*. Because most of Hits@3 and Hits@10 for WN datasets are exceed 98\%, we only report Hits@1.}
	\label{tab:result2}
	\begin{tabular}{llcccc}
		\hline
		& \multirow{2}{*}{Model} & \multicolumn{2}{c}{WN18*} & \multicolumn{2}{c}{WN18RR*} \\
		& & MR(\textit{filter}) & Hits@1 & MR(\textit{filter}) & Hits@1 \\
		\hline
		\multirow{7}{*}{\rotatebox{90}{\textbf{T}}} & TransE & 1.99 & 92.88 & 1.97 & 92.86  \\
		& TransH & 1.73 & 94.04 & 1.70 & 94.40  \\
		& TransR & 1.56 & 94.89 & 1.57 & 95.04  \\
		& CTransR & 1.54 & 95.17 & 1.54 & 95.22  \\
		& DistMult & 1.25 & 96.60 & 1.24 & 96.82  \\
		& ComplEx & 1.26 & 96.64 & 1.22 & 96.69  \\
		& ConvE & 1.17 & 97.23 & 1.13 & 97.17  \\
		& RotatE & 1.16 & 97.31 & 1.10 & 97.31 \\
		\hline
		\hline
		\multirow{4}{*}{\rotatebox{90}{\textbf{T\&P}}} & rTransE & 1.41 & 95.72 & 1.40 & 95.77  \\
		& PTransE(MUL 2-step) & 1.56 & 94.75 & 1.53 & 94.96  \\
		& PTransE(RNN 2-step) & 1.38 & 95.82 & 1.36 & 96.07  \\
		& PTransE(ADD 2-step) & 1.20 & 96.97 & 1.19 & 96.88  \\
		\hline
		\hline
		\multirow{6}{*}{\rotatebox{90}{\textbf{P}}} & B-PR & 1.85 & 93.61 & 1.87 & 93.49   \\
		& SFE-PR & 1.79 & 93.93 & 1.82 & 93.53  \\
		& Ours (Word2Vec\&Shortest) & 1.49 & 95.42 & 1.47 & 95.60 \\
		& Ours (Word2Vec\&RW) & 1.41 & 95.72 & 1.42 & 95.95  \\	
		& Ours (TransE\&Shortest) & 1.18 & 97.15 & 1.15 & 97.15  \\
		& Ours (TransE\&RW) & \textbf{1.13} & \textbf{97.39} & \textbf{1.07} & \textbf{97.44}  \\
		\hline
	\end{tabular}
\end{table}

\begin{table*}[t]
\centering
\caption{Comparison of different feature extraction methods in FB15K.}
\label{tab:featureextractor}
\begin{tabular}{lcccccc}	
	\hline
	Method & Trainable & Linear & MR & Hits@1 & Hits@3 & Hits@10 \\		
	\hline
	Non-Linear Encoder (multi-layers) & Yes & No & \textbf{1.37} & \textbf{95.81} & \textbf{97.78} & \textbf{98.37}\\
	Non-Linear Encoder (single layer) & Yes & No & 2.13 & 92.65 & 94.24 & 95.64 \\
	\hline
	\hline
	Principal Component Analysis & No & Yes & 4.68 & 78.47 & 80.33 & 80.95 \\
	Linear Discriminant Analysis & No & Yes & 4.57 & 79.06 & 80.89 & 81.54 \\
	Locally Linear Embedding & No & No & 4.05 & 81.92 & 83.72 & 84.26 \\
	\hline
\end{tabular}
\end{table*}

To investigate the performance of our model in detail, we analyze the performance of our model with different relation frequency, as shown in Figure~\ref{fig:bucket}.
Since the overall distribution of relations in the five datasets clearly shows a long tail phenomenon (the statistics of single relations on datasets are shown in Figure~\ref{fig:distribution}), we divide the relations into three buckets~\cite{xie2017acl}.
With each bucket, we compare our method (TransE\&RW) with TransE and PTransE (ADD 2-step).
The results show that introducing path information can improve the classification effect in all buckets, and our method is significantly better than TransE and PTransE(ADD 2-step) in dealing with low-frequency relations.

Combining the analysis performed over KBs and the statistic information in Table~\ref{tab:dataset}, we can conclude that the denser (the ratio of entities to relations) and the larger (the amount of entities and relations) the KB becomes, the better our method outperforms baselines.
This means that our method is better suited for relation completion of large-scale and dense KBs.

\subsubsection{Parameter Study}
In order to measure the impact of fluctuations in the hyperparameters of our model on KBC performance, we perform the following experimental analysis for different hyperparameters.

\begin{figure}[t]
	\centering
	\includegraphics[width=1\linewidth]{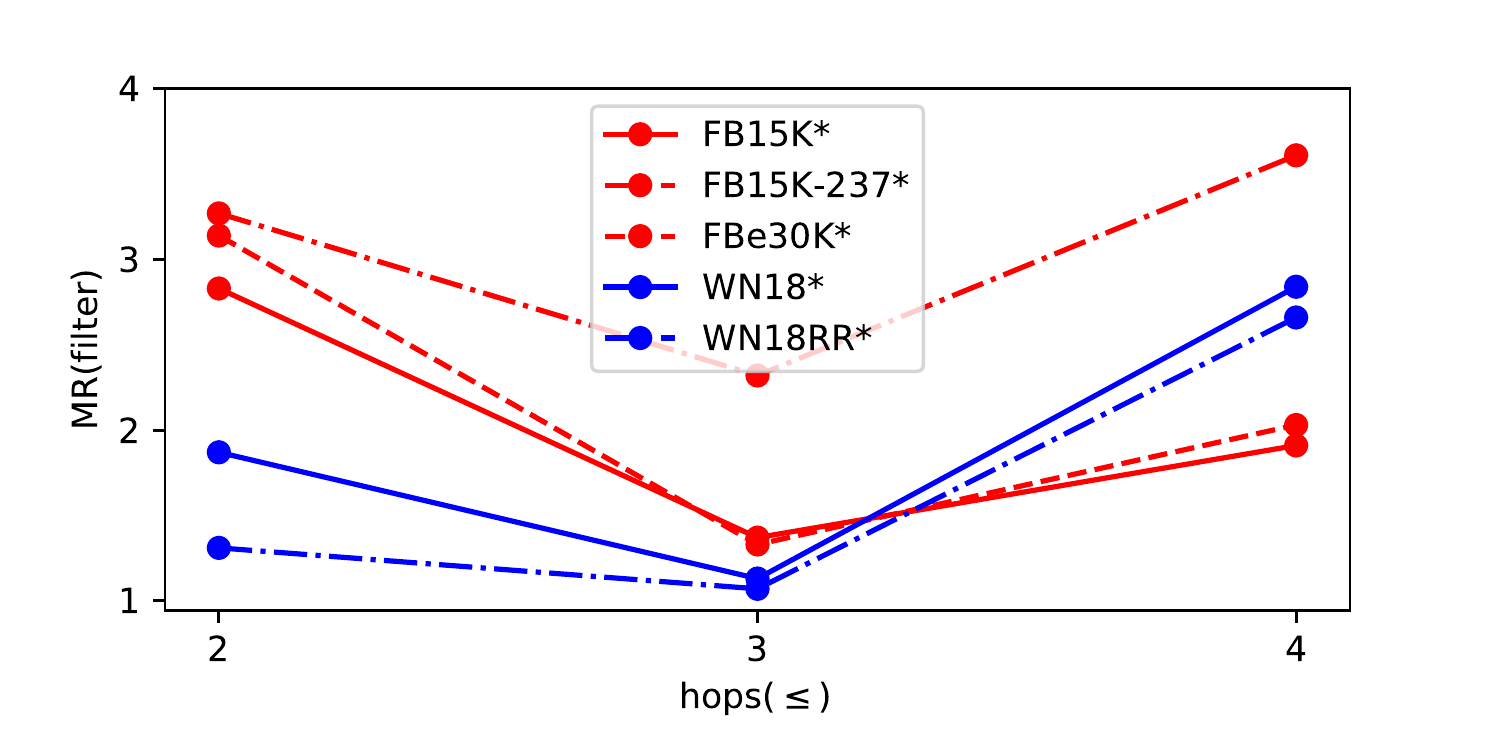}
	\caption{Comparison of different hop upper limit selection. The lower the MR(filter), the better.}\label{fig:hops}
\end{figure}

\textit{Path Length}: We set the maximum path length (i.e. maximum hop limit) of the selected multi-hop paths between the paired entities in the range of 2 to 4 to observe the impact of path length on model performance.
The computational complexity dramatically raises with the increase of maximum path length allowed, so we stop at 4 which still takes acceptable computational time on our reduced datasets.
The experimental results shown in Figure~\ref{fig:hops} demonstrate that setting the parameter to 3 achieves significantly better results that setting it to 2.
However, as we allow longer path to be sampled (e.g. length of 4), prediction accuracy is not further improved, instead it is weakened.
We believe that this is because as the maximum path length increases, the number of sampled multi-hop paths increases rapidly, among which the proportion of noisy/useless paths also grows greatly.
Through observations we have noticed that a 5-hop path can almost connect any two entities in our datasets, which definitely introduces a large amount of noise.
We also noticed that there are only 18 kinds of relations in the WN, which results in a higher sensitivity to changes in path length limitations, as shown in Figure~\ref{fig:hops}.

\begin{figure}[t]
	\centering
	\includegraphics[width=1\linewidth]{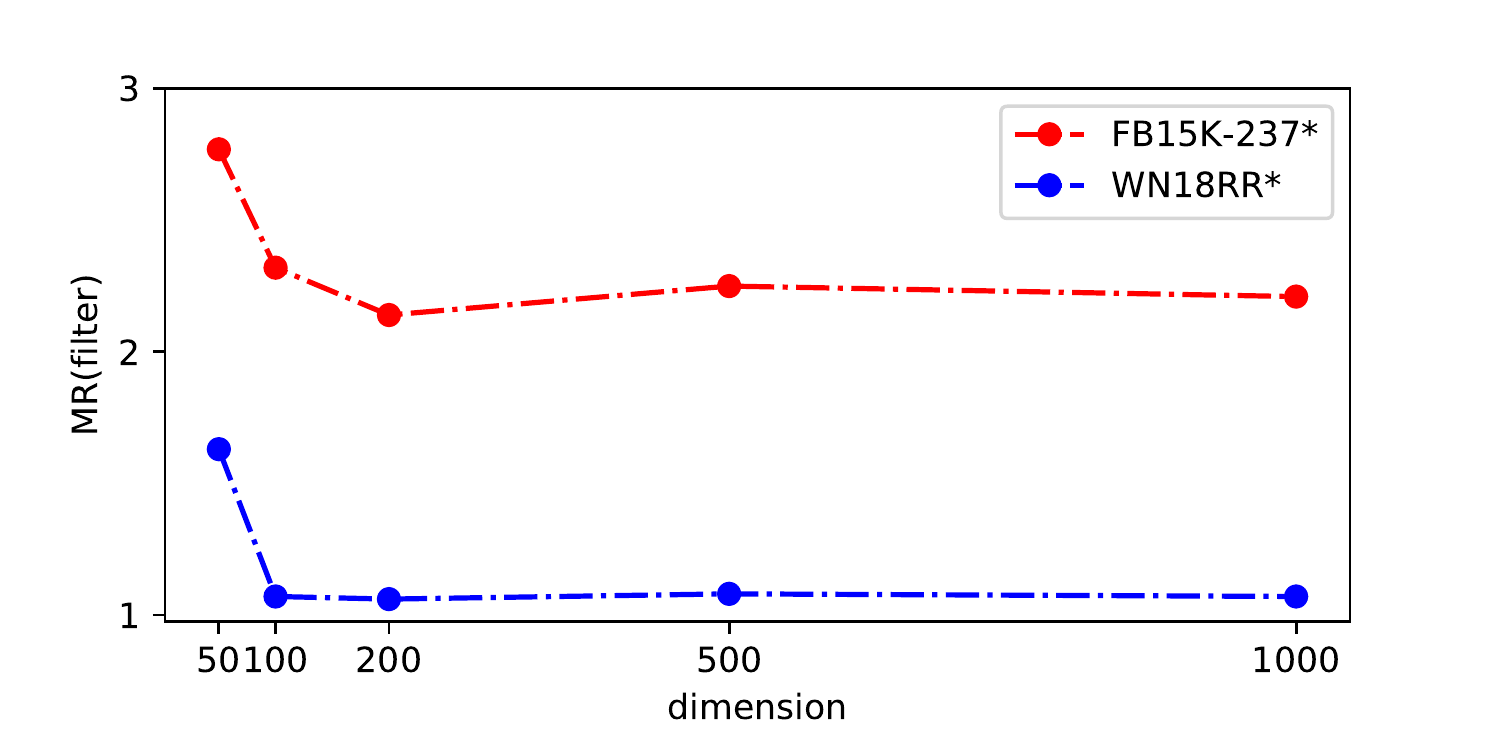}
	\caption{Comparison of different knowledge embedding dimension. The lower the MR(filter), the better.}\label{fig:dimension}
\end{figure}

\textit{Embedding Dimension}:
As a hyperparameter of KB embedding, the dimension of the embeddings may influence the semantic information contained in the input relations and paths, which may then bring variance to our model.
To verify this, we set the embedding dimension to 50, 100, 200, 500, and 1000 respectively and test our model on FB15k-237 and WN18RR\footnote{The varying dimension does not include the position embedding and direction embedding.}.
The experimental results in Figure~\ref{fig:dimension} illustrate that the model is sensitive to the variance of embedding dimension when it is low, and becomes insensitive to the increasing dimension as it increases.
In particular, the inflection point of the WN's curve takes place earlier than that of the FB'S, which we believe is because fewer relations in FB lets it require fewer dimensions to distinguish different semantic information between the relations.

\begin{figure}[t]
	\begin{minipage}[t]{0.98\linewidth}
		\label{fig:var}
		\centering
		\includegraphics[width=1\textwidth]{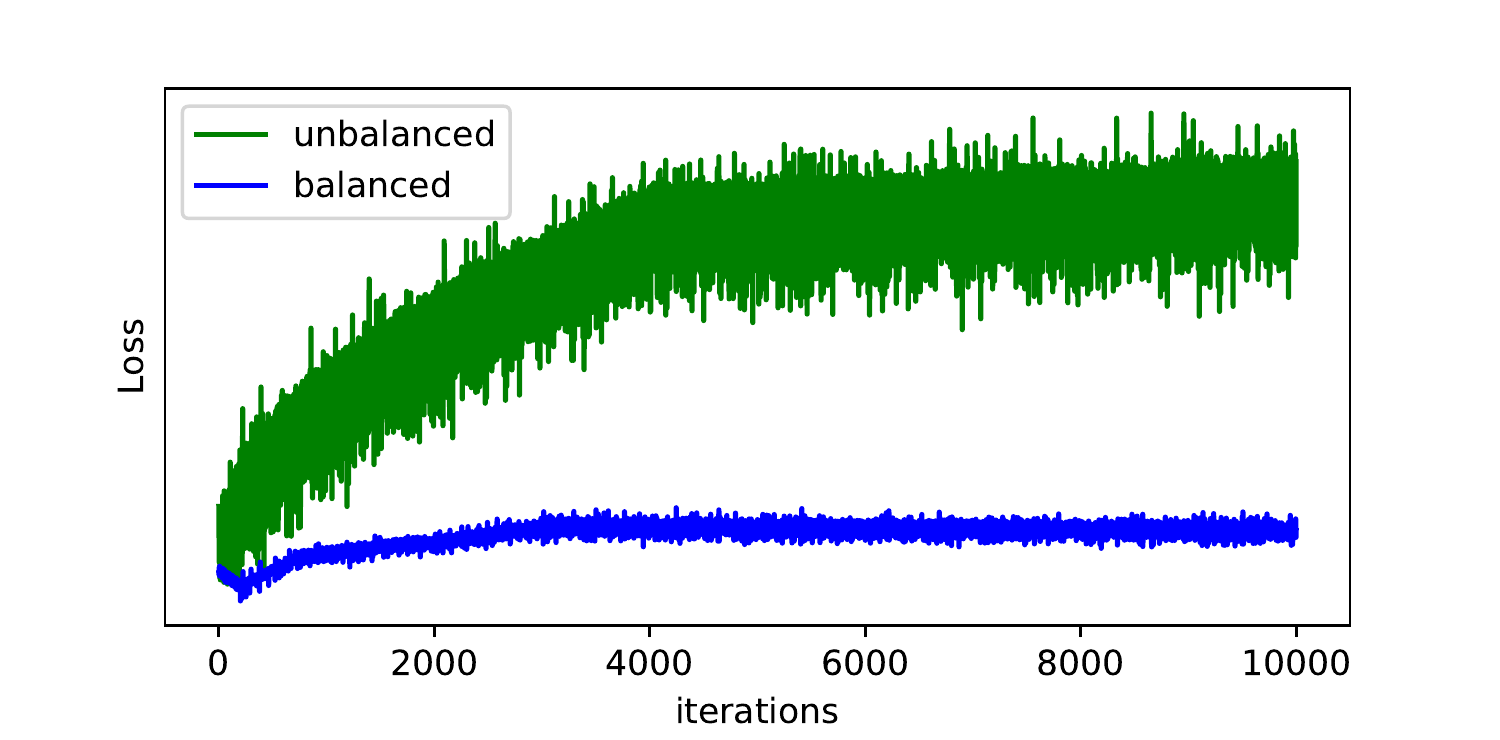}
		\centerline{Loss of each iteration.}
	\end{minipage}
	\begin{minipage}[t]{0.98\linewidth}
		\label{fig:var1}
		\centering
		\includegraphics[width=1\textwidth]{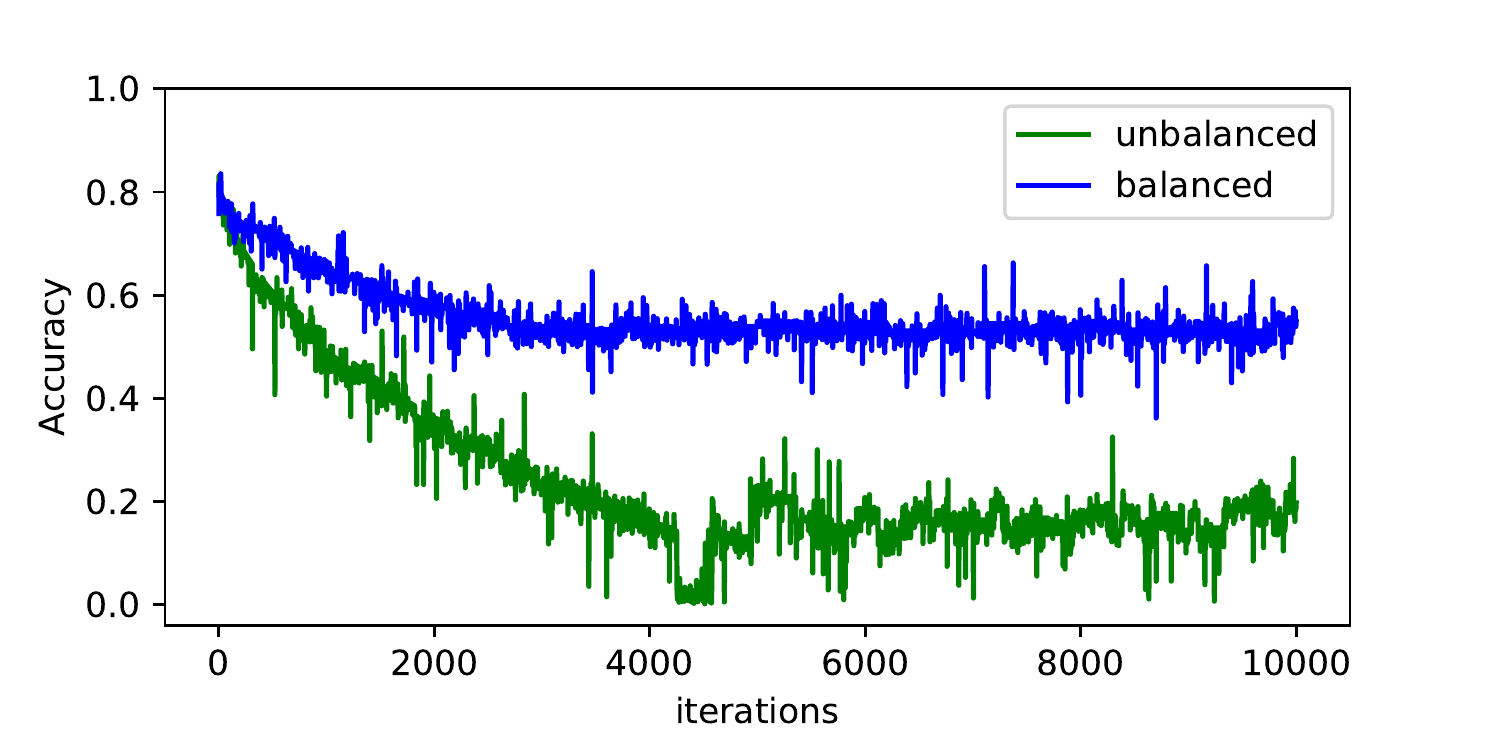}
		\centerline{Classification accuracy.}
	\end{minipage}
	\caption{The loss and classification accuracy (with GRL) training on the FB15k* varies with iteration.}
	\label{fig:bal}
\end{figure}

\textit{Data Balance}:
The source discriminator in our model plays an important role in ensuring that the features extracted by the feature extractor are the shared ones between the single relations and multi-hop paths.
However, the training process is hindered by uneven distribution of the two sources as the number of multi-hop paths greatly exceeds that of the single relations.
To overcome this, we manually adjust the distribution of samples from the two sources in each mini-batch (of size 236) to balance the proportion.
The loss of the source discriminator through iterations before and after the data balance adjustments are shown in Figure~\ref{fig:bal}.
The results show that the source discriminator (fed with balanced data) can eventually meet the expected classification accuracy of approximately 0.5, indicating that the source can not be distinguished.
Notice that data balance also exists in the relation classifier to tackle similar situations in learning the low-frequency relations.

\subsubsection{Model Interpretability}
To validate that our model can select the informative multi-hop paths and crucial relations in them, we choose sample triples and visualize the intermediate results of the hierarchical attention layers in Table~\ref{tab:attention}, Table~\ref{tab:attention1}, and Table~\ref{tab:attention2} as a case study.
The single relations in the tables that should be complemented are: \texttt{/location/location/people\_born\_here}, \texttt{/locati\\on/country/languages\_spoken}, and \texttt{/people/pers\\on/education./education/education/institut\\ion}.
By observing the weights of different paths and the relations in one path, we can find that the crucial paths and relations are given higher attention weights.
For example, as illustrated in Figure~\ref{tab:attention}, relations \texttt{place\_of\_birth}$^{-1}$, \texttt{place\_lived/l\\ocation}$^{-1}$, and \texttt{nationality}$^{-1}$ descent in importance for determining the missing relation \texttt{people\_born\_here}.
Correspondingly, paths containing them are given decreasing weights.
All three relations are given relatively higher weights in their paths, yet their weights decline with decreasing importance.
The above results indicate that the HANs in our model can effectively determine the importance of paths and encode them into low dimensional vector by organically combining the semantic information contained in the composing relations.

\begin{figure}[t]
	\centering
	\includegraphics[width=1\linewidth]{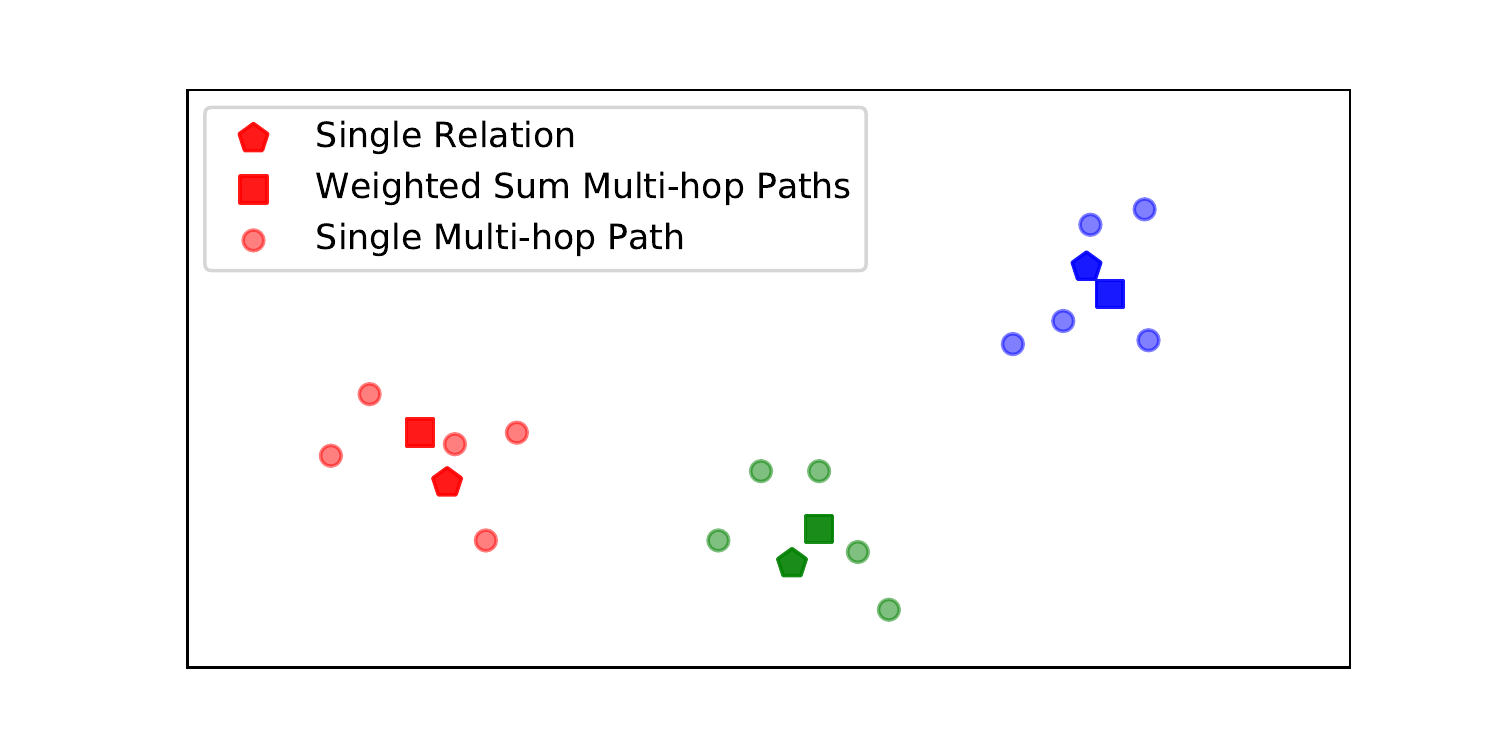}
	\caption{The visualization of dimension reduction results of the single relation $r$ in FB15K*, single path codings $p_{i}$ (top 5) and weighted sum codings $p$. Three colors represent three different relations respectively.}
	\label{fig:reduce}
\end{figure}

To further demonstrate this, we randomly select three different single relations and subsequently the most important multi-hop paths that decide them in FB15K*.
We use PCA to reduce the dimensions of their embeddings and visualize the result in Figure \ref{fig:reduce}, which shows that the weighted sum of the multi=hop paths is closer to the missing single relation than each multi-hop path alone, indicating a semantically closer relationship between them.
It effectively verifies that the HANs can effectively incorporate the semantic information of the multi-hop paths.

\begin{figure}[t]
	\begin{minipage}[t]{0.49\linewidth}
		\label{fig:var}
		\centering
		\includegraphics[width=1\textwidth]{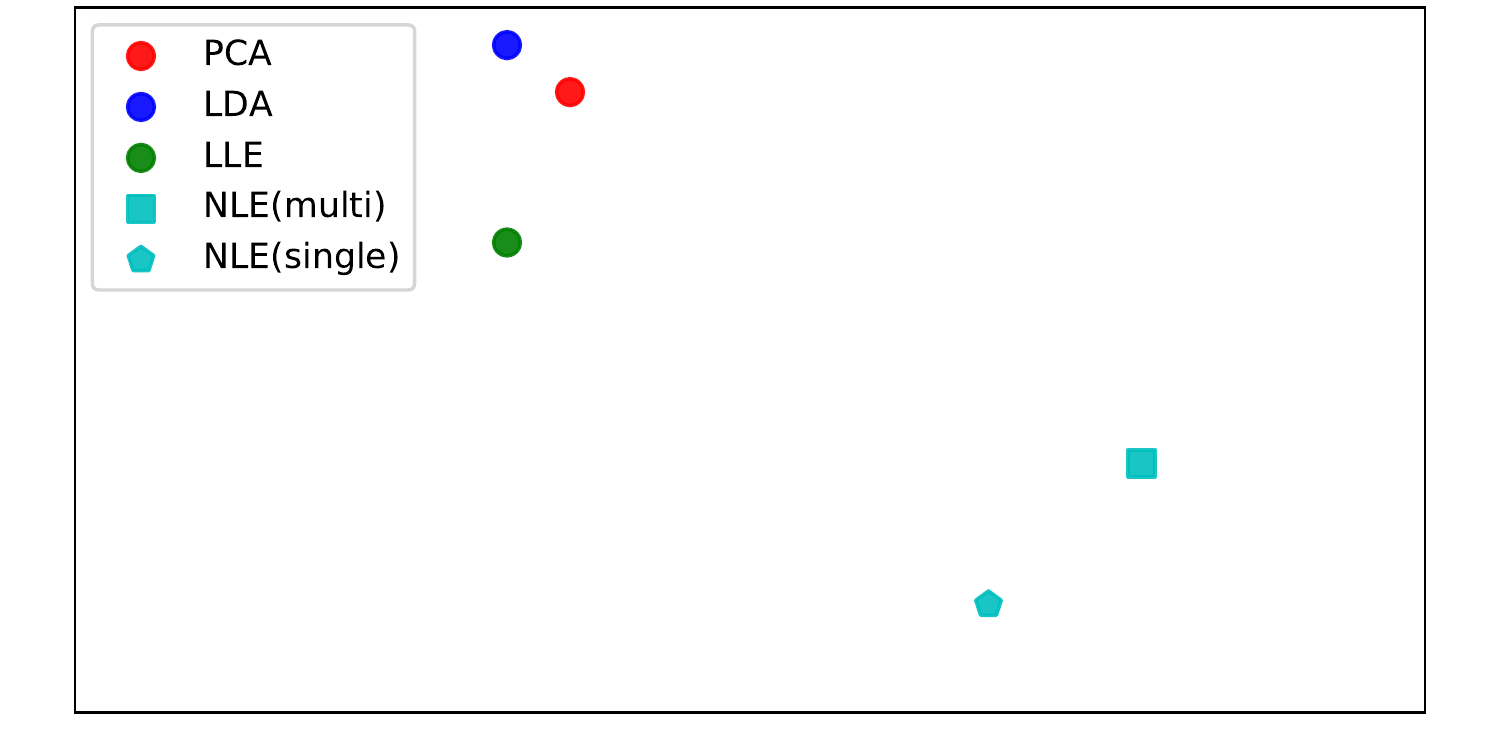}
		\centerline{1. \small{\texttt{../founders}}}
	\end{minipage}
	\begin{minipage}[t]{0.49\linewidth}
		\label{fig:var1}
		\centering
		\includegraphics[width=1\textwidth]{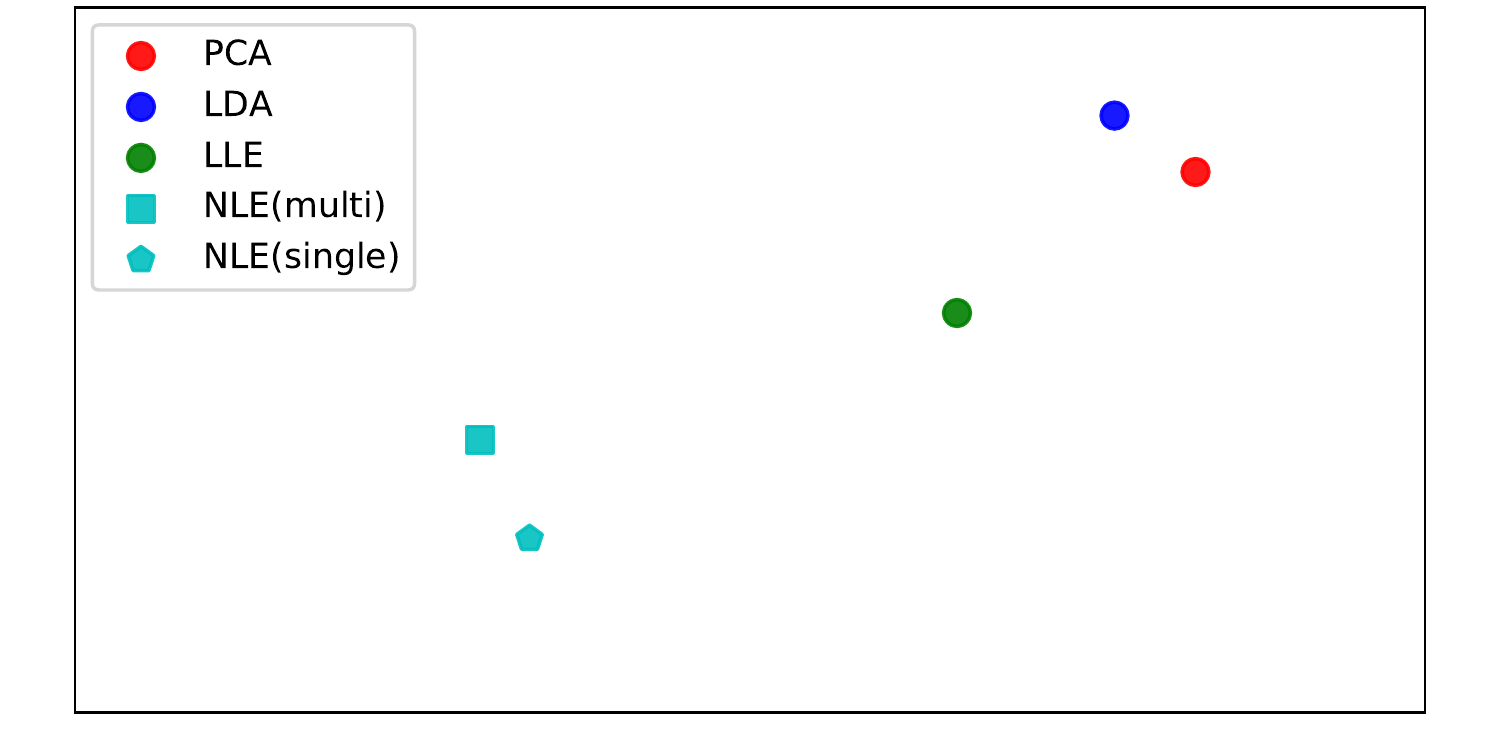}
		\centerline{2. \small{\texttt{../\_government}}}
	\end{minipage}
	\begin{minipage}[t]{0.49\linewidth}
		\label{fig:var1}
		\centering
		\includegraphics[width=1\textwidth]{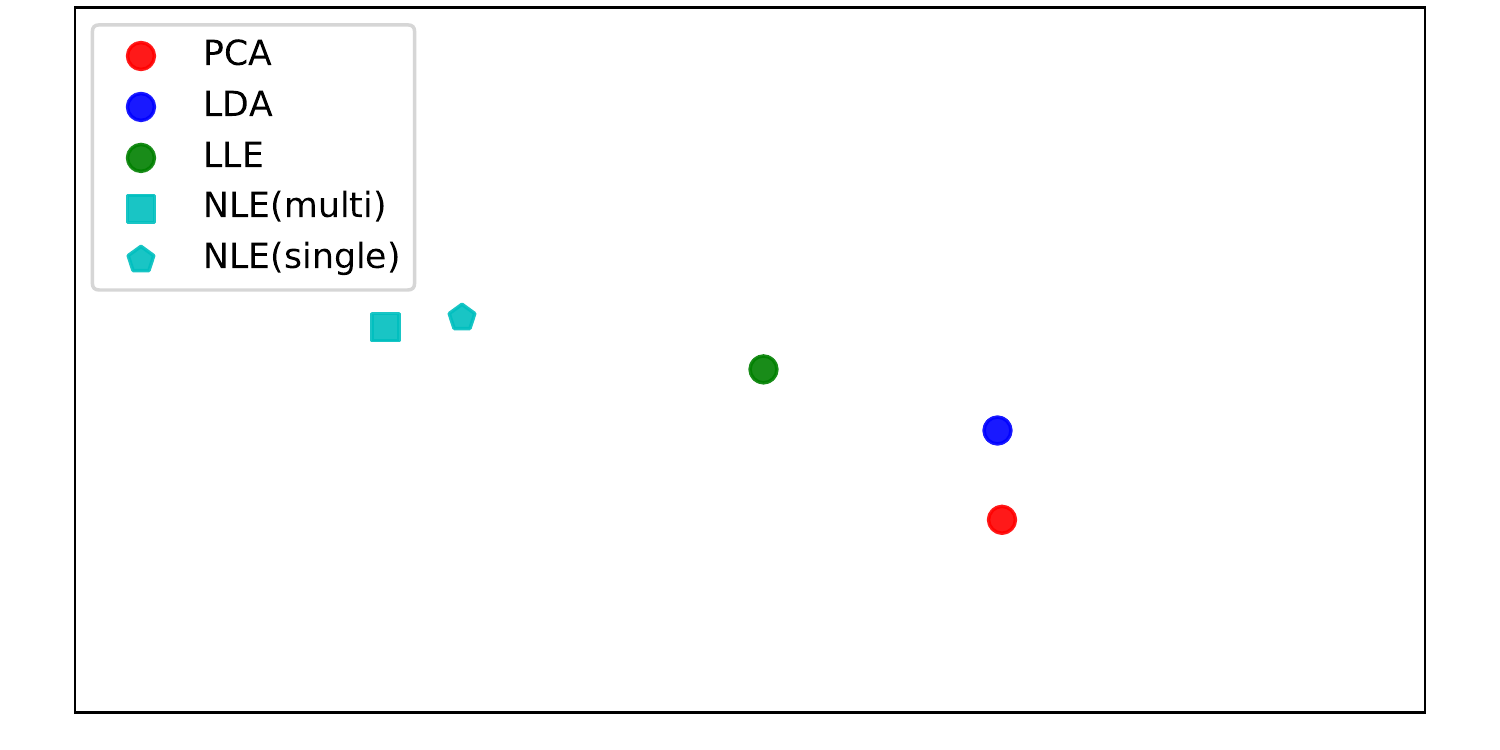}
		\centerline{3. \small{\texttt{../\_family}}}
	\end{minipage}
	\begin{minipage}[t]{0.49\linewidth}
		\label{fig:var1}
		\centering
		\includegraphics[width=1\textwidth]{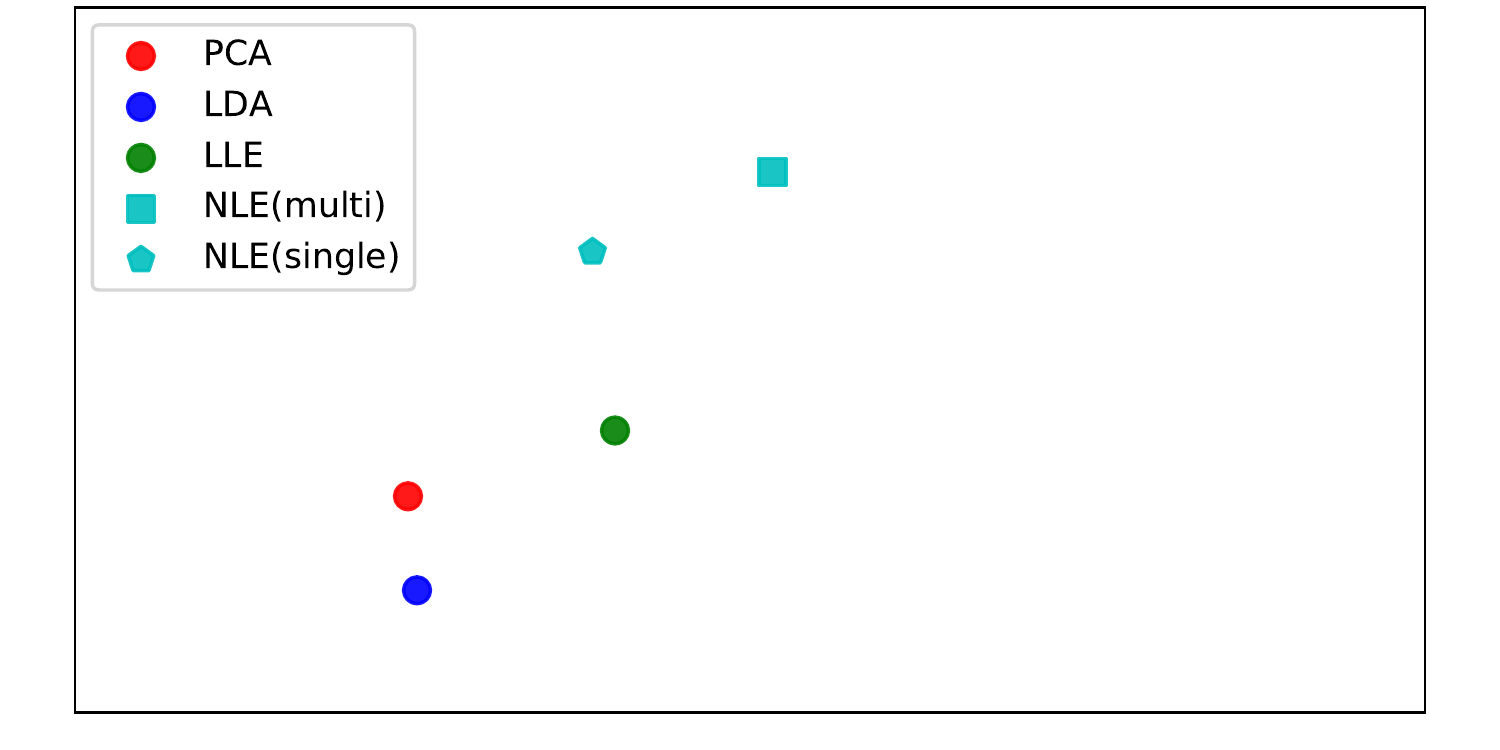}
		\centerline{4. \small{\texttt{../language}}}
	\end{minipage}
	\caption{
		The visualization of examples in FB15K* for dimension reduction using different algorithms.
		We find that in the vast majority of dimension reduction results, the outputs of our feature extractor clearly deviate from other methods (which retain original information as much as possible).
	}
	\label{fig:relreduce}
\end{figure}

We compare our feature extractor with several common dimension reduction algorithms on FB15K*~\cite{Martinez2002PCA,marzia2001nips}.
Experimental results illustrated in Table~\ref{tab:featureextractor} show that our method has a higher classification performance.
To further analyze the difference between our method and dimension reduction algorithms, we visualize the features extracted with different methods in a common vector space as shown in Figure~\ref{fig:relreduce}.
We find that our feature extractor, which is driven by the downstream sub-modules of the proposed model, gets completely different results from the traditional dimension reduction methods (i.e. PCA, LDA, and LLE), which aim at retaining information of the original data as much as possible, while the results of the latter do not show much variance.
This indicates that the upper-stream HANs has reduced and discarded some of the noisy/useless information for downstream tasks, which coincides with our description in Section~\ref{per}.
Fundamentally, the visualization results validate our assumption to some extent, i.e. there exists shared semantic information and noisy/useless information between the single relation and multi-hop paths.

\begin{table*}[h]
	\scriptsize
	\centering
	\caption{The weight assignment of different relation paths in FB15K. The order of the columns indicates the ranking of the weights of path attention (descending order), and the color depth of background in each path indicates the importance of relation attention (the deeper the more important).}
	\vspace{0pt}
	\label{tab:attention}
	\begin{tabular}{|p{1cm}<{\centering}|p{14cm}<{\raggedright}|}
		\hline
		Rank & \hspace{0.4\linewidth}Multi-hop Path \\
		\hline
		0 & \texttt{/location/location/people\_born\_here} \\
		\hline
		1 & \colorbox{llg}{\texttt{/location/statistical\_region/rent50\_2./measurement\_unit/dated\_money\_value/currency}} $\rightarrow$ \colorbox{dg}{\texttt{/location/location/people\_born\_here}} \\
		2 & \colorbox{llg}{\texttt{/location/statistical\_region/rent50\_2./measurement\_unit/dated\_money\_value/currency}} $\rightarrow$ \colorbox{dg}{\texttt{/people/person/place\_of\_birth}$^{-1}$}\\
		3 & \colorbox{llg}{\texttt{/location/statistical\_region/rent50\_2./measurement\_unit/dated\_money\_value/currency}} $\rightarrow$ \colorbox{dg}{\texttt{/people/person/places\_lived./people/place\_lived/location}$^{-1}$} \\
		4 & \colorbox{llg}{\texttt{/tv/tv\_series\_episode/guest\_stars./tv/tv\_guest\_role/actor}} $\rightarrow$ \colorbox{lg}{\texttt{/location/hud\_county\_place/place}}  $\rightarrow$ \colorbox{dg}{\texttt{/location/location/people\_born\_here}} \\
		5 & \colorbox{llg}{\texttt{/tv/tv\_series\_episode/guest\_stars./tv/tv\_guest\_role/actor}} $\rightarrow$ \colorbox{lg}{\texttt{/base/biblioness/bibs\_location/country}} $\rightarrow$ \colorbox{dg}{\texttt{/people/person/nationality}$^{-1}$} \\
		\hline
	\end{tabular}
\end{table*}
\begin{table*}[h]
	\scriptsize
	\centering
	\caption{The visualization result of \texttt{/location/country/languages\_spoken}.}
	\label{tab:attention1}
	\begin{tabular}{|p{1cm}<{\centering}|p{14cm}<{\raggedright}|}
		\hline
		Rank & \hspace{0.4\linewidth}Multi-hop Path \\
		\hline
		0 & \texttt{/location/country/languages\_spoken} \\
		\hline
		1 & \colorbox{llg}{\texttt{/location/statistical\_region/rent50\_2./measurement\_unit/dated\_money\_value/currency}} $\rightarrow$ \colorbox{dg}{\texttt{/location/country/languages\_spoken}} \\
		2 & \colorbox{llg}{\texttt{/location/statistical\_region/rent50\_2./measurement\_unit/dated\_money\_value/currency}} $\rightarrow$ \colorbox{dg}{\texttt{/language/human\_language/countries\_spoken\_in}} \\
		3 & \colorbox{llg}{\texttt{/tv/tv\_series\_episode/guest\_stars./tv/tv\_guest\_role/actor}} $\rightarrow$ \colorbox{dg}{\texttt{/location/country/languages\_spoken}$^{-1}$} $\rightarrow$ \colorbox{lg}{\texttt{/location/location/adjoin\_s./location/adjoining\_relationship/adjoins}} \\
		4 & \colorbox{llg}{\texttt{/location/statistical\_region/rent50\_2./measurement\_unit/dated\_money\_value/currency}} $\rightarrow$ \colorbox{dg}{\texttt{/location/country/official\_language}} \\
		5 & \colorbox{llg}{\texttt{/tv/tv\_series\_episode/guest\_stars./tv/tv\_guest\_role/actor}} $\rightarrow$ \colorbox{lg}{\texttt{/base/locations/continents/countries\_within}$^{-1}$} $\rightarrow$ \colorbox{dg}{\texttt{/language/human\_language/region}$^{-1}$} \\
		\hline
	\end{tabular}
\end{table*}
\begin{table*}[h]
	\scriptsize
	\centering
	\caption{The visualization result of \texttt{/people/person/education./education/education/institution}.}
	\label{tab:attention2}
	\begin{tabular}{|p{1cm}<{\centering}|p{14cm}<{\raggedright}|}
		\hline
		Rank & \hspace{0.4\linewidth}Multi-hop Path \\
		\hline
		0 & \texttt{/people/person/education./education/education/institution} \\
		\hline
		1 & \colorbox{llg}{\texttt{/location/statistical\_region/rent50\_2./measurement\_unit/dated\_money\_value/currency}} $\rightarrow$ \colorbox{dg}{\texttt{/people/person/education./education/education/institution}} \\
		2 & \colorbox{llg}{\texttt{/location/statistical\_region/rent50\_2./measurement\_unit/dated\_money\_value/currency}} $\rightarrow$ \colorbox{dg}{\texttt{/education/educational\_institution/students\_graduates./education/education/student}$^{-1}$} \\
		3 & \colorbox{llg}{\texttt{/tv/tv\_series\_episode/guest\_stars./tv/tv\_guest\_role/actor}} $\rightarrow$ \colorbox{dg}{\texttt{/people/person/education./education/education/institution}} $\rightarrow$ \colorbox{dg}{\texttt{/education/educational\_institution\_campus/educational\_institution}} \\
		4 & \colorbox{llg}{\texttt{/tv/tv\_series\_episode/guest\_stars./tv/tv\_guest\_role/actor}} $\rightarrow$ \colorbox{dg}{\texttt{/people/person/education./education/education/institution}}  $\rightarrow$ \colorbox{lg}{\texttt{/education/educational\_institution/campuses}$^{-1}$} \\
		5 & \colorbox{llg}{\texttt{/tv/tv\_series\_episode/guest\_stars./tv/tv\_guest\_role/actor}} $\rightarrow$ \colorbox{dg}{\texttt{/people/person/education./education/education/institution}}  $\rightarrow$ \colorbox{llg}{\texttt{/dataworld/gardening\_hint/split\_to}$^{-1}$} \\
		\hline
	\end{tabular}
\end{table*}

\section{Conclusion}\label{s5}
In this paper, we propose a novel method for encoding and extracting the shared features between the single relation and multi-hop paths among paired entities based on the simple hypothesis that the connecting relation and paths generally share a vast amount of semantic information.
We utilize it for KBC without any manual pivots.
Experiments on five knowledge base datasets demonstrate the effectiveness of our model.
Furthermore, by independently verifying each sub-module, we show that through AT, the HANs and coupled neural networks in our model has successfully captured the indistinguishable yet valuable features and used it to complete the missing relations.
In the future, we will explore the potential of our model in other artificial intelligence areas, e.g., distant supervised relation extraction.

\section{Acknowledge}
This work is supported by the NSFC program (No. 61772151), and National Key R\&D Program of China (No.2018YFC0830804).

\end{document}